\documentclass{article}

\usepackage[main, final]{neurips_2026}

\usepackage[utf8]{inputenc} 
\usepackage[T1]{fontenc}    
\usepackage{hyperref}       
\usepackage{url}            
\usepackage{booktabs}       
\usepackage{amsfonts}       
\usepackage{amssymb}        
\usepackage{amsmath}        
\usepackage{nicefrac}       
\usepackage{microtype}      
\usepackage{xcolor}         
\usepackage{graphicx}       
\usepackage{placeins}       
\usepackage{float}          
\usepackage{adjustbox}      

\definecolor{darkblue}{rgb}{0, 0, 0.5}
\hypersetup{colorlinks=true, citecolor=darkblue, linkcolor=darkblue, urlcolor=darkblue}

\title{Language models lack systematicity in generating and updating hypotheses}

\author{%
  Hua-Dong Xiong \\
  School of Psychological and Brain Sciences\\
  Georgia Tech\\
  \texttt{hdx@gatech.edu} \\
}

\begin{document}

\maketitle

\begin{abstract}

Large language models (LLMs) are increasingly applied to complex problems, from automated code debugging to science itself. Such problems require generating plausible hypotheses from sparse, noisy, and ambiguous data and updating them as more information arrives. Yet how closely LLMs approach optimal inductive reasoning remains unclear. We study this question in the classic number game, in which a learner infers a concept from a few ambiguous positive integers: $\{2, 16, 4, 8\}$, for example, is explained both by ``powers of 2'' and by ``numbers up to 20.'' We probe the concepts LLMs hypothesize from the observed examples and use their behavioral patterns to understand how they generate and update hypotheses. We measure the LLM posterior over hypotheses in three tasks (prediction, evaluation, and generation) and compare LLM behavior with a Bayesian model and with humans. LLMs favor narrower hypotheses than humans and show a robust evaluation--generation gap: in evaluation they select hypotheses more consistent with the observed examples, whereas in generation they propose narrower hypotheses that are more often rules. They also update poorly: as examples accumulate, their accuracy declines while their confidence grows. Thinking reduces, and can even reverse, the gap; it also broadens the hypotheses LLMs favor, raising accuracy while lowering confidence. Whereas prior studies have shown that LLMs are good hypothesis proposers, our results and further comparisons with the Bayesian model show that LLM inductive reasoning lacks coherence across the three tasks, suggesting that LLMs do not maintain a stable representation shared across tasks that make the same demand. Our results highlight a limitation of LLMs in inductive inference, particularly in settings such as science, where hypotheses must go beyond the data.

\end{abstract}

\section{Introduction}

Imagine fine-tuning a large language model (LLM) for a downstream task. You vibe-code a pipeline, the training loss drops, and you are happy. But on test generations, nothing improves. You quickly form plausible hypotheses: too many epochs, a wrong mask, a train/eval mismatch. After a night of tests, you narrow the possibilities and blame low-quality datasets collected by your colleagues.

Humans are naturally good scientists: from sparse, noisy, and ambiguous data, we form hypotheses about how the world could be, use them to predict what comes next, and update them as new data arrive \citep{vonhelmholtz_facts_1878, tenenbaum_how_2011}. Children see only a few dogs yet extend the word ``dog'' to fluffy animals with four legs, but do not extend the word ``mom'' in the same way \citep{clark_whats_1973, macnamara_cognitive_1972, rescorla_overextension_1980}. Mendeleev inferred periodic structure from limited, noisy measurements, and that hypothesis generalized to new elements. This inductive inference from a few observations supports both everyday problem solving and scientific progress \citep{lake_humanlevel_2015}.

As LLMs grow more capable, they increasingly carry out tasks autonomously, from agentic coding to scientific research and even the improvement of AI systems themselves, known as recursive self-improvement \citep{zhang_darwin_2026, oh_discovering_2025, zweiger_selfadapting_2025, karpathy_autoresearch_2026}. To solve these problems, LLMs propose hypotheses, run analyses, and update their hypotheses with data \citep{battleday_artificial_2024, cornelio_combining_2023, gottweis_ai_2025, yamada_ai_2025, lu_endtoend_2026, mitchener_kosmos_2025, novikov_alphaevolve_2025, murphy_model_2026, kasenberg_aidiscovered_2026, jagadish_closing_2026}. Previous work shows that LLMs propose good hypotheses yet often fail to apply them consistently \citep{qiu_phenomenal_2024}, but it remains unclear whether LLMs hold the right hypothesis and merely fail to execute it, or lack a single representation of the hypothesis shared by proposing and applying it.

Because the quality of the complex decisions that LLMs now make autonomously rests on inductive inference, we need to understand how LLMs generate and update hypotheses, where they fail, and what those failures imply. In this paper, we investigate these questions in a setting whose hypothesis space is simple to parameterize and analyze: the number game \citep{tenenbaum_rules_1999, tenenbaum_bayesian_1999}. A learner observes a few positive integers and infers the concept that generated them. After seeing $\{2, 16, 4, 8\}$, for example, one might consider rule hypotheses such as powers of 2 or even numbers, or interval hypotheses such as numbers from 1 to 20. Because this handful of examples fits many hypotheses, the hypotheses a learner chooses reveal its inductive biases. Since the hypothesis space is small enough to enumerate, we can use a Bayesian model to characterize LLM behavior and compare LLMs against it. Humans favor the narrowest hypothesis that contains the examples \citep{tenenbaum_rules_1999, xu_word_2007}: for $\{2, 16, 4, 8\}$, they prefer powers of 2 over the much larger set of even numbers. This preference, the size principle, follows from assuming that the examples were drawn from the concept \citep{xu_sensitivity_2007, navarro_sampling_2012}: if the concept were all even numbers, seeing only powers of 2 would be a coincidence that becomes exponentially less likely with each new example.

\begin{figure}[htbp]
  \centering
  \includegraphics[width=\linewidth]{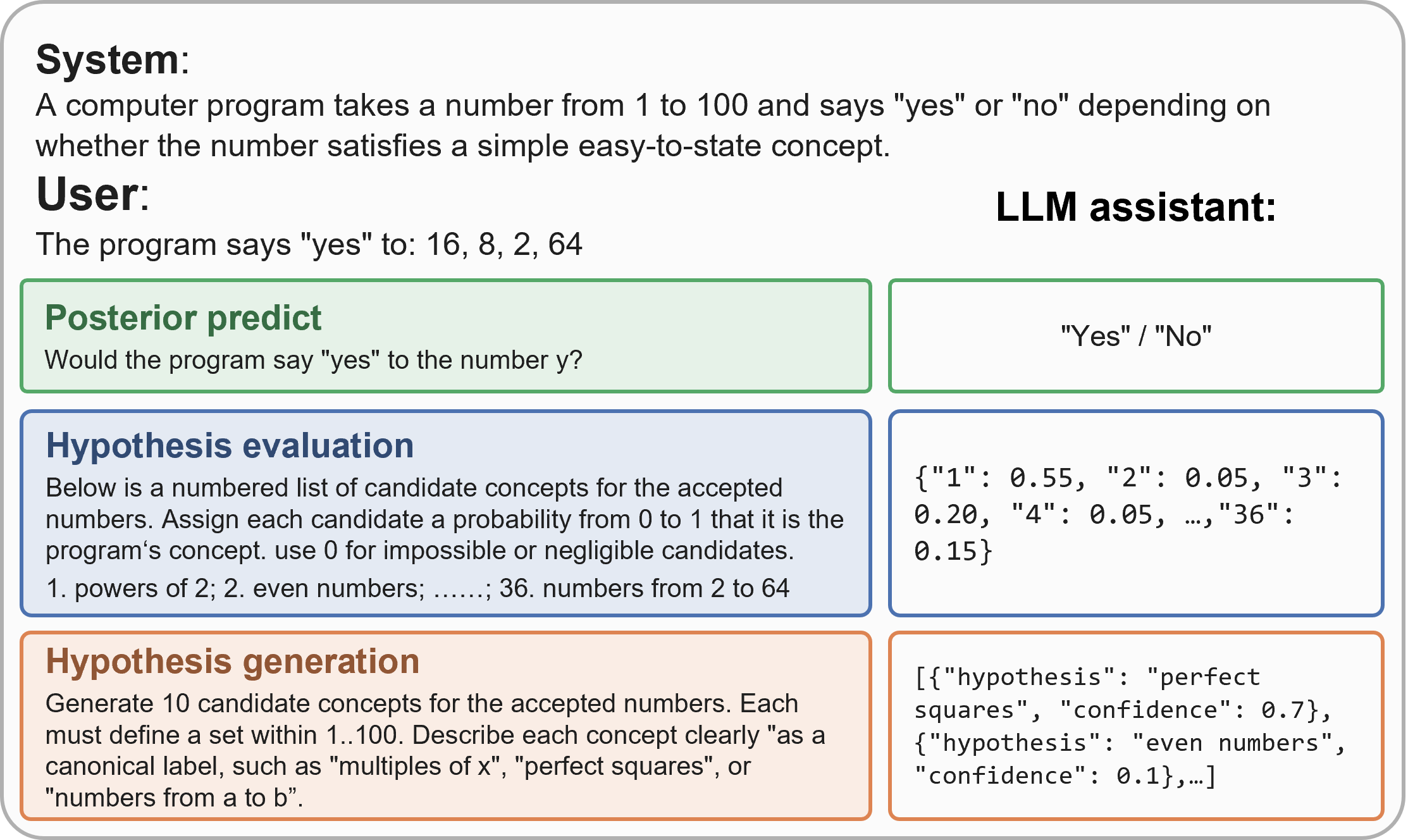}
  \caption{\textbf{Three measurements of the posterior over hypotheses in the number game.} The schematic shows how we prompt LLMs to measure their posterior over hypotheses. Posterior prediction queries the LLM with one integer in the integer domain at a time and records the model-generated probability that the integer belongs to the same concept as the current examples, yielding a membership probability for every integer in the domain. Hypothesis evaluation shows the candidate hypothesis list, the rule hypotheses and a few interval hypotheses around the current examples drawn from the hypothesis space of the Bayesian model, and records the LLM's confidence in each listed hypothesis. Hypothesis generation asks the LLM to generate 10 different hypotheses and associated confidences given the examples.}
  \label{fig:tasks}
\end{figure}

We measure the same inductive inference in three ways (Fig.~\ref{fig:tasks}). Posterior prediction asks whether a new number belongs with the examples; hypothesis evaluation shows the model a candidate hypothesis list and asks how much it believes each hypothesis; hypothesis generation asks it to propose hypotheses that explain the examples. If a learner applies the same inferential principle across problems that share the same structure, the three measurements should agree. This property, systematicity \citep{fodor_connectionism_1988}, has long been debated for neural networks \citep{lake_generalization_2018, lake_humanlike_2023}. Previous work has found a generator--validator gap in LLMs: they often verify answers better than they generate them \citep{cobbe_training_2021, li_benchmarking_2023, west_generative_2023, rodriguez_rankalign_2025}. In inductive inference, such a gap would be a failure of systematicity, and because the Bayesian model specifies what each measurement should show, the number game lets us measure where and why the hypotheses LLMs generate depart from those they evaluate.

We first use the Bayesian model to describe the inductive biases LLMs bring to forming hypotheses from sparse and ambiguous data. LLMs favor narrow hypotheses more strongly than humans, that is, they apply a stronger size principle, especially when examples are few, in non-thinking mode, and in smaller models. Second, consistent with previous work on the generator--validator gap, our three measurements yield different behaviors: evaluation commits to hypotheses that fit the examples better and with more confidence, whereas generation proposes narrower ones. Thinking narrows this gap and even reverses it in accuracy. Third, as LLMs update their hypotheses over accumulating examples, they keep choosing narrow hypotheses even at the cost of accuracy, and they grow more confident without becoming more accurate; thinking improves the accuracy of their hypotheses but leaves these inductive biases largely unchanged. Fourth, we extend the integer domain to $\{1,\ldots,200\}$ while LLMs see only examples from $\{1,\ldots,100\}$. When the examples suggest an interval, the size principle that LLMs apply so strongly within the original domain should withdraw belief from $101..200$ as examples accumulate, yet most LLMs stop applying it there: thinking makes this worse for prediction and evaluation, and in generation nearly every model fails to apply it. Finally, prompts that state how the examples were sampled or that show the candidate hypothesis list leave the size principle largely intact. LLMs thus do not seem to turn a description of the generative process into a model they can reason with.

Together, these results show that LLM inductive inference lacks systematicity: the same models apply different size principles depending on how their beliefs are measured and whether their hypotheses must extend beyond the observed examples, and their size principle departs from the Bayesian model by changing with the number of examples. This limitation matters most for scientific inference, where hypotheses must go beyond the data, and it points to opportunities for making future models more Bayesian \citep{qiu_bayesian_2026} or for pairing them with symbolic models that structure and calibrate their inferences \citep{liu_cognitive_2026, murphy_model_2026}. More broadly, judging inductive inference in LLMs requires asking whether one representation of a hypothesis serves prediction, evaluation, generation, and extension beyond the observed examples alike, and understanding why the representations of current LLMs lack this systematicity.

\section{Methods}

\subsection{Tenenbaum's number game}
\label{sec:methods_game}

The number game was introduced as a minimal setting for studying how people reason inductively \citep{tenenbaum_rules_1999, tenenbaum_bayesian_1999}. Given examples such as $\{2, 16, 4, 8\}$, a learner might infer a rule hypothesis such as powers of 2, a broader rule such as even numbers, or an interval hypothesis around the observed examples. Because a few positive examples are compatible with many hypotheses, the task lets us measure the inductive biases that shape hypothesis generation and updating.

Formally, the hypothesis space $\mathcal{H}$ is a finite set of candidate hypotheses. Each hypothesis $h\in\mathcal{H}$ denotes a subset of a finite integer domain $D_d=\{1,\ldots,d\}$, and learning consists of inferring which subset generated the observed examples. The original number-game experiments used the finite domain $\{1,\ldots,100\}$; we use the same domain and also test the extended domain $\{1,\ldots,200\}$ to measure LLM behavior beyond the range from which the examples are drawn. The hypothesis space combines two intuitive families: rule hypotheses, which follow mathematical patterns (e.g., even numbers or powers of 2), and interval hypotheses, which cover contiguous ranges of magnitude (e.g., numbers from 10 to 20). Following \citet{tenenbaum_rules_1999, tenenbaum_bayesian_1999}, the prior assigns mass across these rule and interval hypotheses; construction details and prior parameters are given in Appendix~\ref{sec:appendix_candidates}. The candidate hypothesis list shown in hypothesis evaluation is drawn from the same rule-and-interval hypothesis space.

The likelihood is determined by the assumed sampling process for positive examples. Under strong sampling, examples are assumed to be drawn uniformly from inside the true hypothesis; under weak sampling, they are only known to be positive instances. Strong sampling yields the \emph{size principle}: observing several examples inside a narrow hypothesis is more informative than observing the same examples inside a broad hypothesis. If $X$ is the observed examples and $|h|$ is the number of domain elements in $h$, the strong-sampling likelihood is proportional to
\begin{equation}
  p(X \mid h) \propto |h|^{-|X|}
  \quad \text{if } X \subseteq h,
\end{equation}
and zero otherwise. Thus four examples consistent with powers of 2 are much more likely under that narrow rule than under a broad hypothesis such as even numbers.
This is the implicit Occam's razor of strong sampling: among hypotheses that explain the examples, narrower compatible hypotheses receive more posterior support.

The Bayesian model predicts whether a new number $y$ belongs to the concept by averaging over hypotheses, $p(y\mid X)=\sum_{h\in\mathcal{H}}\mathbf{1}[y\in h]\,p(h\mid X)$.

To compare LLM behavior to this model, we fit a version of the Bayesian model in which the prior over hypotheses, the pooled prior, is learned jointly from all models and human participants, and only the strength of the size principle, an exponent $\beta$ on the likelihood above, is estimated separately for each model, measurement, and number of examples; we call the result the size-principle fit. A size principle of 1 recovers the strong-sampling likelihood of the Bayesian model, larger values penalize broad compatible hypotheses more strongly, and 0 removes the penalty; because the prior is shared, size-principle values are comparable across models, numbers of examples, and measurements. The pooled prior, the estimation procedure, and a comparison of alternative parameterizations are given in Appendices~\ref{sec:appendix_abfit} and~\ref{sec:appendix_prior_comparison}.

\subsection{Probing LLMs' posterior over hypotheses via three measurements}

We measure LLMs' posterior over hypotheses given examples $X$ in three ways. Posterior prediction queries every integer $y$ in the integer domain $D_d$ and records the probability $q(y\mid X)$ that the model answers Yes to a question about $y$ (Appendix~\ref{sec:appendix_elicitation}). Hypothesis evaluation provides the candidate hypothesis list, drawn from the same rule-and-interval construction used by the Bayesian model, and asks for the confidence assigned to each hypothesis (Appendix~\ref{sec:appendix_candidates}). Hypothesis generation asks the model to propose 10 hypotheses that describe the observed examples, with an associated confidence for each hypothesis. Under the Bayesian model, these three measurements should agree, because all three derive from one posterior $p(h\mid X)$. We compare them by projecting the confidences of evaluation and generation into the same predictive space as posterior prediction and fitting the size principle to all three measurements in that space. This projection lets us ask whether the measurement itself changes the hypotheses the model generates and updates. Details of the candidate hypothesis list and the projection are given in Appendices~\ref{sec:appendix_candidates} and~\ref{sec:appendix_hypmetrics}.

\subsection{Stimuli and models}
\label{sec:methods_stimuli}

We use stimulus sets and human data from two datasets: \textsc{Tenenbaum99}, which contains eight hand-designed stimulus sets from the classic number-game experiments, including rule-like sets such as $\{16, 8, 2, 64\}$ and similarity-like sets such as $\{16, 23, 19, 20\}$, and \textsc{Bigelow16}, which contains 255 stimulus sets from the broader Bigelow and Piantadosi human dataset \citep{bigelow_large_2016,bigelow_inferring_2016}. LLMs see each stimulus set one example at a time in context, and we measure their posterior after each of the first one to four examples; the examples shown so far are the observed examples $X$. This gives 26 combinations of stimulus set and number of examples per measurement for \textsc{Tenenbaum99} and 636 for \textsc{Bigelow16} in the original domain; Appendix~\ref{sec:appendix_stimuli} gives the stimulus categories, and Appendix~\ref{sec:appendix_abfit} describes how fits pool the two datasets and vary the number of examples.

We evaluate eight LLMs: Gemma 4 A4B and E4B \citep{gemma_2026}; Qwen 3.6 A3B and Qwen 3.5 4B \citep{qwen_qwen35_2026}; GPT-5.6 Luna and GPT-5.4 Mini; Nemotron 3 Nano \citep{nvidia_nvidia_2025}; and Gemini 3.6 Flash. Each model is run in thinking and non-thinking modes. Gemini 3.6 Flash cannot switch thinking off, so its lowest thinking level, \texttt{minimal}, serves as non-thinking mode. The backends that serve GPT-5.6 Luna and Gemini 3.6 Flash return no log-probabilities, so for these two models posterior prediction is estimated from 10 sampled responses (Appendix~\ref{sec:appendix_elicitation}). Results for the stronger model group, Gemma 4 A4B, Qwen 3.6 A3B, GPT-5.6 Luna, and Gemini 3.6 Flash, are given in Appendix~\ref{sec:appendix_best}. Prompt conditions and further experimental details are given in Appendices~\ref{sec:appendix_models_prompts}--\ref{sec:appendix_domain}.

\section{Results}

\subsection{LLMs favor narrow hypotheses most when examples are few}
\label{sec:res_fit}

We first ask what inductive bias LLMs bring to forming hypotheses when the examples are sparse and ambiguous. We quantify this bias with the size principle of the Bayesian model, fit for each model, measurement, and number of examples and compared with the value fitted to human judgments. All fits share the pooled prior, which we learn from the data because the hand-set prior is arbitrary and a prior fit separately for each model trades off with the size principle, leaving it unstable (Appendix~\ref{sec:appendix_abfit}).

\begin{figure}[htbp]
  \centering
  \includegraphics[width=\linewidth]{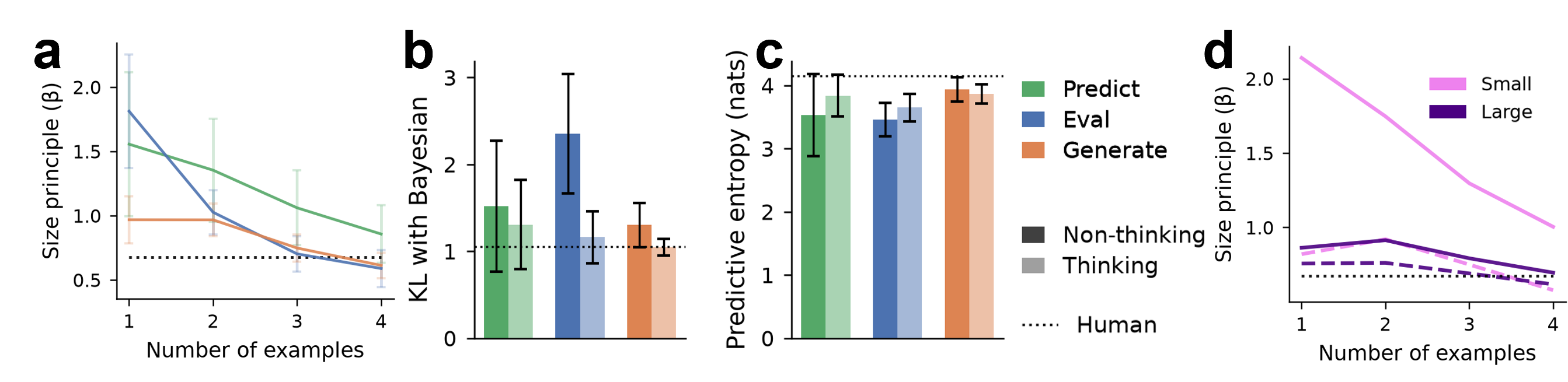}
  \caption{\textbf{Size principle in LLMs.} \textbf{a}, Size principle over one to four examples for posterior prediction (Predict), hypothesis evaluation (Eval), and hypothesis generation (Generate) in non-thinking mode under the \emph{default} prompt, fit after projecting each measurement into the same posterior predictive space; error bars show 95\% confidence intervals across models. \textbf{b,c}, KL divergence from the Bayesian posterior (\textbf{b}) and predictive entropy (\textbf{c}) for each measurement under the \emph{default} prompt. Dark bars show non-thinking mode and lighter bars thinking mode; error bars show 95\% confidence intervals across models. \textbf{d}, Posterior-prediction size principle for smaller (Gemma 4 E4B, Qwen 3.5 4B) and larger (Gemma 4 A4B, Qwen 3.6 A3B) models, averaged within each size; solid and dashed lines show non-thinking and thinking mode. In all panels, the dotted line marks the human value.}
  \label{fig:behaviors}
\end{figure}

After a single example, all three measurements apply a size principle above the human level, and it weakens as examples accumulate (Fig.~\ref{fig:behaviors}a; individual models in Appendix Fig.~\ref{fig:supp_alpha_beta_thinking}a). LLMs thus favor narrow hypotheses most strongly when the examples constrain them least. The three measurements, however, disagree on how strong this preference is: given the same examples, the same models apply different size principles depending on how their beliefs are measured, whereas a single posterior would yield a single size principle.

In non-thinking mode, evaluation also departs furthest from the Bayesian posterior, with the largest Kullback--Leibler (KL) divergence of the three measurements, and prediction and evaluation spread their beliefs less than humans, with lower predictive entropy (Fig.~\ref{fig:behaviors}b,c). Thinking reduces the KL divergence of evaluation to the level of the other two and raises the predictive entropy of prediction and evaluation toward the human level; it also weakens the posterior-prediction size principle at every number of examples (Appendix Fig.~\ref{fig:supp_alpha_beta_thinking}b). Model scale matters in a similar way. Within the Gemma and Qwen families, smaller models in non-thinking mode apply a much stronger size principle than larger ones, especially after a single example, and thinking brings them onto the trajectory of the larger models (Fig.~\ref{fig:behaviors}d). Thinking and model scale thus both move the preference for narrow hypotheses toward the human level.

\subsection{Evaluation favors accuracy, generation narrow rules}
\label{sec:res_gap}

The size-principle fit summarizes the full distribution over hypotheses. We next ask which hypothesis each measurement selects when it must pick one. We compare the top-1 hypothesis, the hypothesis assigned the largest confidence, in evaluation and generation under the \emph{default} prompt, and describe it by its accuracy, that is, its consistency with all observed examples, the one criterion every correct hypothesis must meet; its normalized confidence; the share of the integer domain it covers (support fraction); and whether it takes a rule form (rule proportion). We also ask whether each measurement favors rules more for rule-like than for similarity-like stimulus sets (rule AUC; 0.5 is chance). Full definitions are given in Appendix~\ref{sec:appendix_hypmetrics}.

\begin{figure}[htbp]
  \centering
  \includegraphics[width=\linewidth]{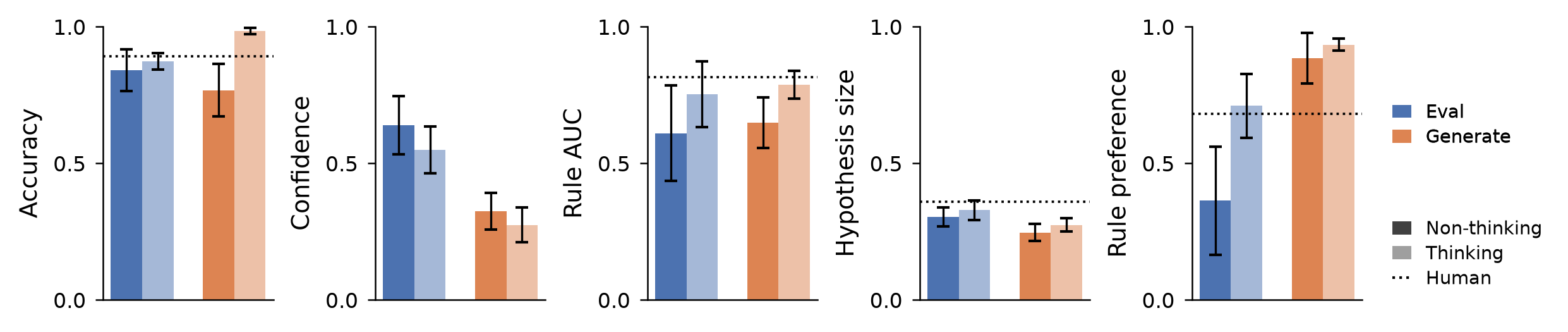}
  \caption{\textbf{Evaluation--generation gap.} Bars compare hypothesis evaluation (Eval) and hypothesis generation (Generate) under the \emph{default} prompt, pooling the two datasets. From left to right, the panels report top-1 accuracy, top-1 normalized confidence, rule AUC, top-1 support fraction (Hypothesis size), and top-1 rule proportion (Rule preference); the top-1 panels average over one to four examples, whereas rule AUC scores each stimulus set with all of its examples. Dark bars show non-thinking mode and lighter bars thinking mode; dotted lines mark values computed from human participants' written descriptions of \textsc{Bigelow16} stimulus sets; error bars show 95\% confidence intervals across models.}
  \label{fig:eval_gen_gap}
\end{figure}

Without thinking, the two measurements commit to different hypotheses (Fig.~\ref{fig:eval_gen_gap}; model-level points in Appendix Fig.~\ref{fig:supp_scatter_rule}a,b). Evaluation selects more accurate top-1 hypotheses and holds them with more confidence, whereas generation proposes narrower hypotheses that more often take a rule form. Across models, a stronger size principle goes with narrower top-1 hypotheses in both measurements (Appendix Fig.~\ref{fig:supp_beta_size}), so the top-1 hypothesis reflects the bias fitted in Section~\ref{sec:res_fit}. Relative to human descriptions, both measurements choose narrower top-1 hypotheses but depart in opposite directions on hypothesis form, generation proposing rules more often than humans and evaluation less often, and neither distinguishes rule-like from similarity-like stimulus sets as well as humans do, evaluation barely above chance. This evaluation--generation gap is the clearest departure from a single Bayesian posterior: given the same examples, the same model commits to different hypotheses depending on whether it judges candidates or proposes its own.

Thinking improves LLM performance on many reasoning tasks \citep{wei_chainofthought_2022, guo_deepseekr1_2025}, and here it raises accuracy as well, so we ask how it changes the hypotheses LLMs commit to. Thinking lifts the accuracy of generation to near ceiling, above that of evaluation, which reverses the gap in accuracy. It also lowers the confidence of both measurements, raises their rule AUC, bringing generation close to humans, and brings the rule proportion of evaluation to the human level (Fig.~\ref{fig:eval_gen_gap}). Although thinking broadens the hypotheses LLMs favor by weakening the size principle (Section~\ref{sec:res_fit}), it barely broadens the top-1 hypotheses they commit to: generation still chooses narrower ones than evaluation, and both stay narrower than humans. Thinking thus raises accuracy and lowers confidence without making the committed hypotheses broader.

\subsection{More examples make LLMs more confident in less accurate hypotheses}
\label{sec:res_update}

A learner that updates well should grow more confident as examples accumulate while keeping its hypotheses consistent with them. We ask whether LLMs update their top-1 hypotheses in this way by tracking the same properties over one to four examples.

\begin{figure}[htbp]
  \centering
  \includegraphics[width=\linewidth]{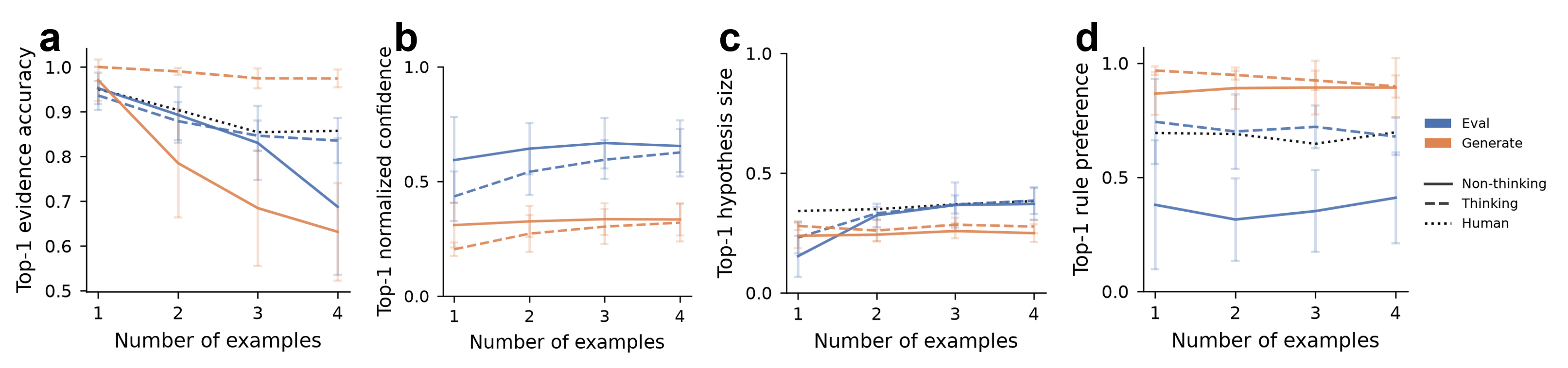}
  \caption{\textbf{Hypothesis updating over examples.} Trajectories over one to four examples of top-1 accuracy (\textbf{a}), normalized confidence (\textbf{b}), support fraction (\textbf{c}), and rule proportion (\textbf{d}) for hypothesis evaluation (Eval) and hypothesis generation (Generate) under the \emph{default} prompt. Solid lines show non-thinking mode and dashed lines thinking mode; the dotted black line marks the value computed from human participants' written descriptions of \textsc{Bigelow16} stimulus sets; error bars show 95\% confidence intervals across models.}
  \label{fig:gen_eval_update}
\end{figure}

LLMs do grow more confident as examples accumulate: evaluation assigns its top-1 hypothesis more confidence with each additional example, and generation holds its confidence (Fig.~\ref{fig:gen_eval_update}b). Their accuracy, however, falls (Fig.~\ref{fig:gen_eval_update}a). Both measurements start near ceiling after one example; without thinking, the accuracy of generation then falls steeply, and that of evaluation falls below the human level by four examples. More examples thus leave LLMs at least as confident in hypotheses that increasingly miss the examples.

The two measurements lose accuracy while updating their hypotheses in different ways. Generation keeps its top-1 hypotheses narrower than humans throughout and favors rule hypotheses more strongly than humans at every number of examples (Fig.~\ref{fig:gen_eval_update}c,d), so each new example that falls outside a narrow rule lowers its accuracy. Evaluation instead broadens its top-1 hypotheses, from well below the human level after one example to the human level by three, in line with its fast-weakening size principle (Section~\ref{sec:res_fit}), and favors rule hypotheses less than humans; its accuracy falls nonetheless.

Thinking removes most of the decline in accuracy, keeping generation near ceiling and evaluation close to humans (Fig.~\ref{fig:gen_eval_update}a), but leaves the other patterns largely in place. Thinking-mode evaluation still grows more confident with each example, thinking-mode generation still keeps narrow rule hypotheses, and only the rule proportion of evaluation moves to the human level (Fig.~\ref{fig:gen_eval_update}b--d). As LLMs update, then, confidence and accuracy come apart, and thinking restores accuracy without changing the narrow rules that generation keeps.

\subsection{LLMs stop applying the size principle beyond the original domain}
\label{sec:res_extension}

The analyses so far stay within the range covered by the observed examples. In scientific inference, however, a hypothesis must also specify how far it reaches beyond the examples that produced it. We therefore extend the integer domain to $\{1,\ldots,200\}$ while all examples stay in $\{1,\ldots,100\}$, and ask whether LLMs apply the size principle to the new range $101..200$.

On similarity-like stimulus sets, no example lies in the new range, so a learner that applies the size principle should shift belief toward hypotheses inside the original domain with each additional example, and the mass it assigns to $101..200$ should fall toward zero, as it does in the Bayesian model. On rule-like stimulus sets, a rule such as powers of 2 extends into the new range, so extension mass should remain. We measure this extension mass for all three measurements on \textsc{Tenenbaum99} stimulus sets, projecting evaluation and generation into the extended predictive space (Appendix~\ref{sec:appendix_domain}).

\begin{figure}[htbp]
  \centering
  \includegraphics[width=\linewidth]{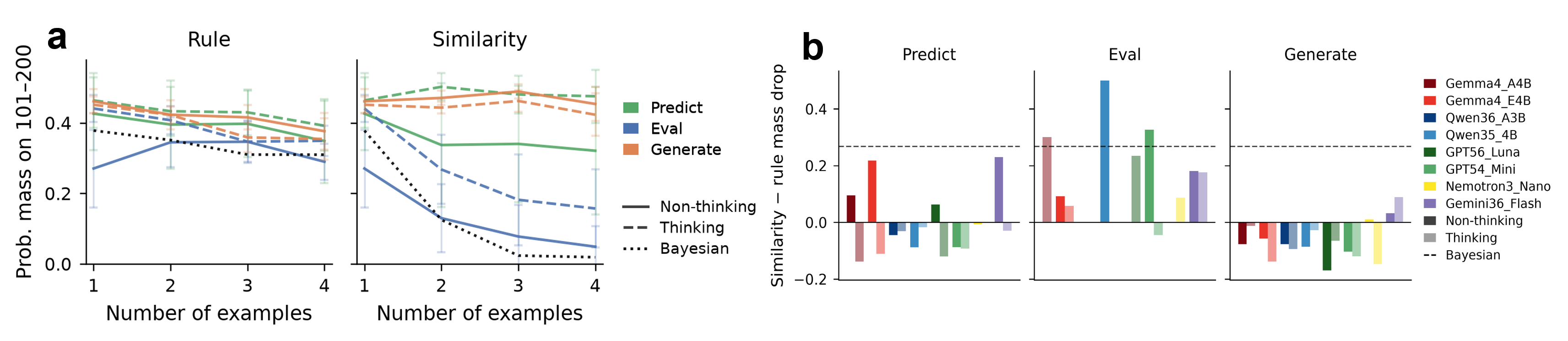}
  \caption{\textbf{LLMs fail to apply the size principle in the new range $101..200$.} \textbf{a}, Mass assigned to $101..200$ in the extended domain as the number of examples grows, for posterior prediction (Predict), hypothesis evaluation (Eval), and hypothesis generation (Generate) on \textsc{Tenenbaum99}, split by rule-like (Rule) and similarity-like (Similarity) stimulus sets. Solid lines show non-thinking mode and dashed lines thinking mode; the Bayesian model is dotted black; error bars show 95\% confidence intervals across models. \textbf{b}, Difference between the decline of extension mass on similarity-like and on rule-like stimulus sets for each model and measurement, where the decline is the mass after one example minus the mean mass after two to four examples. Dark bars show non-thinking mode and lighter bars thinking mode; the dashed line marks the Bayesian model; missing bars mark models whose extension mass is undefined at some number of examples.}
  \label{fig:extension}
\end{figure}

On rule-like stimulus sets, all three measurements assign mass to the new range near the Bayesian model (Fig.~\ref{fig:extension}a). On similarity-like sets, where the Bayesian model withdraws belief from the new range as examples accumulate, most LLM measurements do not. Only non-thinking evaluation follows the Bayesian decline closely; prediction declines after the first example and then stays flat; and generation keeps assigning the new range about as much mass as it does for rule-like sets, so its trajectories barely separate the two kinds of stimulus sets. Thinking makes prediction and evaluation worse, consistent with its weaker size principle within the original domain (Section~\ref{sec:res_fit}): thinking-mode evaluation declines more slowly, and thinking-mode prediction does not decline at all.

Individual models show the same failure. For each model and measurement, we take the difference between the decline of extension mass on similarity-like and on rule-like stimulus sets, where the decline is the mass after one example minus the mean mass after two to four examples; the Bayesian model has a positive difference. Most LLMs fall short of it (Fig.~\ref{fig:extension}b): a few models reach or exceed it in evaluation, none does in prediction, and in generation nearly every model shows a larger decline on rule-like than on similarity-like sets, the reverse of the Bayesian pattern. The failure lies in withdrawing belief rather than in extending rules, since on rule-like sets most models still assign higher posterior-prediction probability to rule-consistent than to rule-inconsistent numbers in the new range (Appendix Fig.~\ref{fig:supp_scatter_rule}c). LLMs thus apply the size principle strongly within the original domain but not beyond it, and generation, which carries its narrow rules into the new range whether or not the examples support a rule, departs from it most.

\subsection{LLMs do not adjust their inductive bias to a stated generative process}
\label{sec:res_prompt}

Under the Bayesian model, the size principle follows from assuming that examples are drawn from the concept and disappears when they are merely positive instances (strong and weak sampling; Section~\ref{sec:methods_game}). A prompt can state this generative process directly. We therefore ask whether LLM posterior prediction changes when the prompt states either assumption (\emph{strong} and \emph{weak} prompts) or shows the candidate hypothesis list (\emph{explicit} prompt), compared with the \emph{default} prompt, which does neither. A learner that adjusts its inductive bias to these descriptions should strengthen its size principle under the \emph{strong} prompt, weaken it under the \emph{weak} prompt, and move toward the Bayesian posterior under the \emph{explicit} prompt. Each prompt condition is fit under its own pooled prior; full prompt definitions are given in Appendix~\ref{sec:appendix_models_prompts}.

\begin{figure}[htbp]
  \centering
  \includegraphics[width=\linewidth]{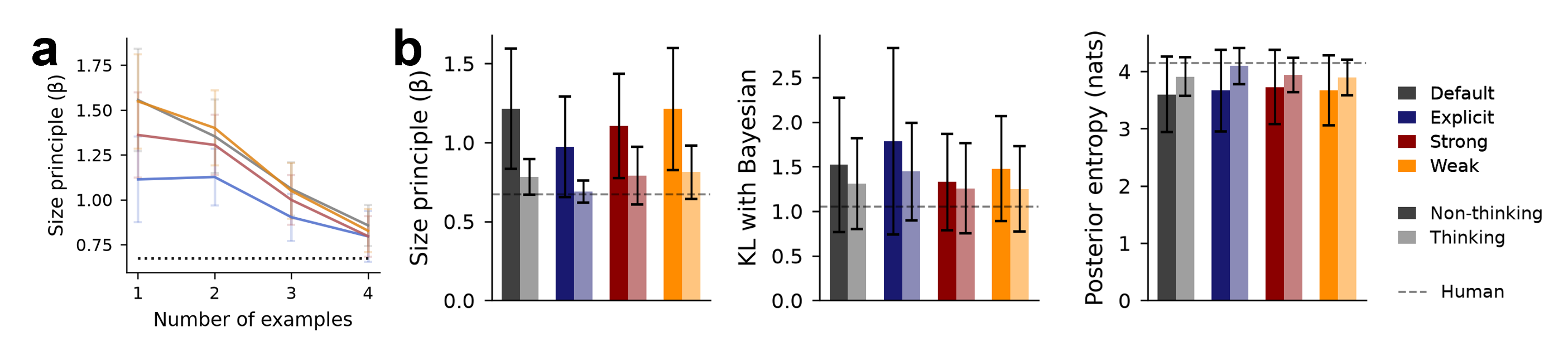}
  \caption{\textbf{Size principle across prompts that describe the generative process.} \textbf{a}, Posterior-prediction size principle over one to four examples for \emph{default}, \emph{explicit}, \emph{strong}, and \emph{weak} prompts in non-thinking mode, each fit under that prompt condition's own pooled prior; the dotted line marks the human value. \textbf{b}, Model-averaged size principle, KL divergence from the Bayesian posterior, and predictive entropy (Posterior entropy) for each prompt condition. Within each prompt condition, dark bars show non-thinking mode and lighter bars thinking mode; dashed lines mark human values. Error bars show 95\% confidence intervals across model-level means.}
  \label{fig:condition}
\end{figure}

Stating the sampling assumption leaves the size principle unchanged. Size principles are similar under \emph{default}, \emph{strong}, and \emph{weak} prompts and all exceed that of humans, so LLMs behave as if examples were drawn from the concept whether or not they are told so (Fig.~\ref{fig:condition}b). The \emph{strong} prompt, which should strengthen the size principle, tends to weaken it instead, most visibly after a single example (Fig.~\ref{fig:condition}a,b). Showing the candidate hypothesis list changes little: it weakens the size principle after a single example, but all four prompt conditions converge by four examples, and the \emph{explicit} prompt does not bring prediction closer to the Bayesian posterior. Thinking moves behavior more than any prompt: within every prompt condition, it weakens the size principle toward the human level and raises predictive entropy while changing KL divergence little (Fig.~\ref{fig:condition}b). Told how the examples were generated, LLMs thus do not change their inductive bias accordingly; their size principle follows a default resembling strong sampling, which thinking can shift but prompts cannot.

\section{Discussion and limitations}

We studied the inductive biases with which LLMs form and update hypotheses, measuring one posterior over hypotheses in three ways and comparing it with the Bayesian model and with humans. The central finding is that LLM inductive inference lacks systematicity: given the same examples, the same model favors different hypotheses depending on how its beliefs are measured. The three measurements apply size principles of different strength (Section~\ref{sec:res_fit}); evaluation selects more accurate hypotheses, whereas generation proposes narrower rule hypotheses (Section~\ref{sec:res_gap}); and in the new range $101..200$, non-thinking evaluation withdraws belief on similarity-like stimulus sets almost as the Bayesian model does, whereas generation carries its rules beyond the original domain whatever the examples suggest (Section~\ref{sec:res_extension}). These results bear on whether LLMs hold the right hypothesis but fail to apply it, or lack one representation shared by proposing and applying it \citep{qiu_phenomenal_2024}, and they favor the latter: a single posterior applied imperfectly would not make the measurements diverge in consistent directions, so LLMs do not seem to maintain a stable representation shared across tasks that make the same demand. Thinking reduces the evaluation--generation gap and reverses it in accuracy, but it does not restore systematicity, because beyond the original domain it moves prediction and evaluation further from the Bayesian model.

Beyond this lack of systematicity, the inductive bias itself departs from the Bayesian model. LLMs apply a stronger size principle than humans when examples are few, and their size principle weakens as examples accumulate, whereas the Bayesian model keeps it fixed (Section~\ref{sec:res_fit}). As examples accumulate, LLMs also grow more confident in top-1 hypotheses that increasingly fail to contain those examples (Section~\ref{sec:res_update}). Finally, prompts that state the sampling assumption or supply the candidate hypothesis list leave the size principle unchanged, whereas thinking weakens it (Section~\ref{sec:res_prompt}). Together, these departures suggest that LLMs bring a default bias resembling strong sampling to every set of examples instead of adapting it to the examples they see or to the generative process they are told.

These failures matter for the larger tasks now delegated to LLMs, such as autonomous research loops in which models propose hypotheses, run analyses, and update on the results \citep{gottweis_ai_2025, lu_endtoend_2026, mitchener_kosmos_2025}. Such loops often pair a model that generates hypotheses with one that verifies them, implicitly treating the two as applying the same beliefs. The evaluation--generation gap shows that they do not: the verifier judges by a different standard from the one by which the generator proposes, and because it can rank only the hypotheses it receives, the coverage of generation bounds what the loop can reach. Updating adds a second risk, since a loop whose confidence rises as its hypotheses fit the data less well can grow more certain while drifting from the data. Scientific hypotheses must also go beyond the data and specify where they stop, and the domain-extension results show that LLMs do not withdraw belief from ranges the examples never reached. Nor can the missing structure simply be stated in the prompt: in terms of amortized inference, a learned mapping from data to beliefs whose cost is paid mainly during learning \citep{gershman_amortized_2014,dasgupta_memory_2021}, an LLM's direct answer is the amortized solution, and redoing the inference for each problem requires the hypothesis space, prior, and likelihood that open-ended scientific problems do not supply.

Several limitations qualify these conclusions. The number game is deliberately small and symbolic: its one-dimensional integer domain and two intuitive hypothesis families make the Bayesian model enumerable, but they are far from the knowledge-rich settings in which people and scientists usually induce hypotheses, so our claims are restricted to inductive tasks of this form. The departures we report are also defined relative to the rule-and-interval hypothesis space and the pooled prior (Appendix~\ref{sec:appendix_candidates}). LLMs may entertain hypotheses outside this space, and the space bounds the fit: even reweighting it freely for each stimulus set fits human judgments below their noise ceiling (Appendix~\ref{sec:appendix_prior_comparison}). Our measurements are behavioral: we infer the posterior from model outputs, so the claim that LLMs lack a stable shared representation is an inference from behavior rather than a direct test of their internal representations. We measured the models' direct inference and did not test scaffolds such as generate-and-verify loops or tools that execute generated hypotheses as code; these could offset some of the failures reported here, but only by moving part of the inference outside the model. Finally, the panel does not include the largest frontier models.

Future work could first test whether the same lack of systematicity recurs in adjacent inductive tasks such as sequence extrapolation, symbolic rule learning, and function induction, and then in knowledge-rich scientific domains. A second direction is to ask why it arises, by testing whether the three measurements draw on a shared internal representation and whether training on one measurement transfers to the others. Thinking also deserves study of its own: analyzing reasoning traces could show whether its mixed effects on systematicity come from a changed search over hypotheses or from a changed bias. The inference could also be made explicit, either by training models toward Bayesian inference \citep{qiu_bayesian_2026} or by pairing LLM hypothesis proposals with symbolic or probabilistic programs that execute and weigh them \citep{ellis_humanlike_2023, wang_hypothesis_2024}, which would test whether systematicity can be recovered outside the model. Finally, our examples were fixed in advance and presented in a fixed order; letting models choose which examples or queries to observe, as in automated experimental design \citep{gandhi_boxinggym_2025, murphy_model_2026}, would test whether they can also gather the data their hypotheses need.

\section{Related work}
\label{sec:related_work}

AI-for-science systems increasingly automate parts of discovery, but the harder step remains proposing, evaluating, and revising the hypotheses that define a problem \citep{battleday_artificial_2024, cornelio_combining_2023, novikov_alphaevolve_2025}, which requires treating hypotheses as revisable objects \citep{popper_logic_2005,lakatos_falsification_2014}. Many such systems close this loop with the model itself, generating candidates and then critiquing, verifying, or refining them \citep{madaan_selfrefine_2023,huang_large_2024}. Their gains depend on a generator--validator gap, the extent to which a model judges outputs better than it produces them \citep{song_mind_2025}. The evaluation--generation gap we report is an instance of this gap, measured in a task where the Bayesian posterior is known.

Most evaluations of inductive inference in LLMs score how accurately models label unseen examples after input--output prompting, which conflates inducing a rule with applying it. Studies that elicit hypotheses directly find that LLMs propose good hypotheses, especially when a program or symbolic interpreter checks and refines them, yet often fail to apply their own rules \citep{wang_hypothesis_2024, qiu_phenomenal_2024}, and that their apparent induction often rests on similarity to nearby observed examples rather than on the induced rule \citep{li_mirage_2025}. Benchmarks extend these tests to rule discovery and scientific hypothesis generation \citep{bertolazzi_falsifybench_2026, liu_hypobench_2026}, and hybrid approaches let LLMs propose natural-language hypotheses for Bayesian inference to weigh \citep{ellis_humanlike_2023, piriyakulkij_doing_2024} or train LLMs to approximate Bayesian or inductive inference directly \citep{qiu_bayesian_2026, zhang_using_2026}. Closest to our work, recent studies test LLMs on probabilistic reasoning, suspicious coincidences, and number-game-like generalization \citep{padmanabhan_language_2025, pournemat_reasoning_2025, bazigaran_concept_2025}, where within a single measurement and a fixed integer domain LLMs can appear approximately Bayesian. These studies score hypotheses and predictions against a single target concept; we compare all three measurements with one Bayesian posterior, which tests whether the hypotheses an LLM generates, the hypotheses it evaluates highly, and the predictions it makes cohere.

Classical accounts of induction are Bayesian: a learner starts from a hypothesis space, computes a posterior with Bayes' rule, and updates it as observations arrive \citep{tenenbaum_bayesian_1999, tenenbaum_how_2011, lake_humanlevel_2015}. The number game makes this account tractable: its rule and interval hypotheses can be enumerated \citep{tenenbaum_rules_1999,tenenbaum_bayesian_1999,bigelow_inferring_2016,bigelow_large_2016}, so the Bayesian model can serve as the standard against which LLMs are compared. Presented in a prompt, the task is also a case of in-context learning, in which examples must change a model's beliefs without parameter updates \citep{brown_language_2020}. Analyses that cast in-context learning as implicit Bayesian inference \citep{xie_explanation_2022, muller_transformers_2021} mostly study small transformers trained on synthetic task families; we instead test LLMs on a classic hypothesis-learning task in which each example incrementally constrains a known posterior.

\begin{ack}
The author gratefully acknowledges support from the Tinker Research Grant and the Center of Excellence in Computational Cognition Pilot Grant, both awarded to H.-D.~X. The author declares no competing interests.
\end{ack}

\section*{Code availability}
\label{sec:code_availability}
The code for reproducing the experiments and analyses in this work is available at \url{https://github.com/sakimarquis/number-game-hypothesis}.

\clearpage

\bibliographystyle{apalike}
\bibliography{references}

@article{lu_endtoend_2026,
	title = {Towards end-to-end automation of {AI} research},
	volume = {651},
	copyright = {2026 The Author(s)},
	issn = {1476-4687},
	url = {https://www.nature.com/articles/s41586-026-10265-5},
	doi = {10.1038/s41586-026-10265-5},
	language = {en},
	number = {8107},
	urldate = {2026-11-06},
	journal = {Nature},
	publisher = {Nature Publishing Group},
	author = {Lu, Chris and Lu, Cong and Lange, Robert Tjarko and Yamada, Yutaro and Hu, Shengran and Foerster, Jakob and Ha, David and Clune, Jeff},
	month = mar,
	year = {2026},
	pages = {914--919},
}

@inproceedings{piriyakulkij_doing_2024,
	title = {Doing {Experiments} and {Revising} {Rules} with {Natural} {Language} and {Probabilistic} {Reasoning}},
	volume = {37},
	url = {https://proceedings.neurips.cc/paper_files/paper/2024/hash/5f1b350fc0c2affd56f465faa36be343-Abstract-Conference.html},
	doi = {10.52202/079017-1682},
	urldate = {2026-11-06},
	booktitle = {Advances in {Neural} {Information} {Processing} {Systems}},
	publisher = {Curran Associates, Inc.},
	author = {Piriyakulkij, Wasu Top and Langenfeld, Cassidy and Le, Tuan and Ellis, Kevin},
	year = {2024},
	pages = {53102--53137},
}

@article{li_mirage_2025,
	title = {{MIRAGE}: {Evaluating} and {Explaining} {Inductive} {Reasoning} {Process} in {Language} {Models}},
	volume = {2025},
	shorttitle = {{MIRAGE}},
	url = {https://proceedings.iclr.cc/paper_files/paper/2025/hash/b782a3462ee9d566291cff148333ea9b-Abstract-Conference.html},
	language = {en},
	urldate = {2026-11-06},
	journal = {International Conference on Learning Representations},
	author = {Li, Jiachun and Cao, Pengfei and Jin, Zhuoran and Chen, Yubo and Liu, Kang and Zhao, Jun},
	month = may,
	year = {2025},
	pages = {73944--73969},
}

@article{wang_hypothesis_2024,
	title = {Hypothesis {Search}: {Inductive} {Reasoning} with {Language} {Models}},
	volume = {2024},
	shorttitle = {Hypothesis {Search}},
	url = {https://proceedings.iclr.cc/paper_files/paper/2024/hash/a934282496e7b907f5d48d49bb4c9d7d-Abstract-Conference.html},
	language = {en},
	urldate = {2026-11-06},
	journal = {International Conference on Learning Representations},
	author = {Wang, Ruocheng and Zelikman, Eric and Poesia, Gabriel and Pu, Yewen and Haber, Nick and Goodman, Noah},
	month = may,
	year = {2024},
	pages = {38993--39014},
}

@misc{karpathy_autoresearch_2026,
	title = {autoresearch: {AI} agents running research on single-{GPU} nanochat training automatically},
	url = {https://github.com/karpathy/autoresearch},
	urldate = {2026-11-06},
	author = {Karpathy, Andrej},
	year = {2026},
}

@inproceedings{west_generative_2023,
	title = {The {Generative} {AI} {Paradox}: “{What} {It} {Can} {Create}, {It} {May} {Not} {Understand}”},
	shorttitle = {The {Generative} {AI} {Paradox}},
	url = {https://openreview.net/forum?id=CF8H8MS5P8},
	language = {en},
	urldate = {2026-11-06},
	author = {West, Peter and Lu, Ximing and Dziri, Nouha and Brahman, Faeze and Li, Linjie and Hwang, Jena D. and Jiang, Liwei and Fisher, Jillian and Ravichander, Abhilasha and Chandu, Khyathi and Newman, Benjamin and Koh, Pang Wei and Ettinger, Allyson and Choi, Yejin},
	month = oct,
	year = {2023},
}

@inproceedings{li_benchmarking_2023,
	title = {Benchmarking and {Improving} {Generator}-{Validator} {Consistency} of {Language} {Models}},
	url = {https://openreview.net/forum?id=phBS6YpTzC},
	language = {en},
	urldate = {2026-11-06},
	author = {Li, Xiang Lisa and Shrivastava, Vaishnavi and Li, Siyan and Hashimoto, Tatsunori and Liang, Percy},
	month = oct,
	year = {2023},
}

@misc{cobbe_training_2021,
	title = {Training {Verifiers} to {Solve} {Math} {Word} {Problems}},
	url = {http://arxiv.org/abs/2110.14168},
	doi = {10.48550/arXiv.2110.14168},
	urldate = {2026-11-06},
	publisher = {arXiv},
	author = {Cobbe, Karl and Kosaraju, Vineet and Bavarian, Mohammad and Chen, Mark and Jun, Heewoo and Kaiser, Lukasz and Plappert, Matthias and Tworek, Jerry and Hilton, Jacob and Nakano, Reiichiro and Hesse, Christopher and Schulman, John},
	month = nov,
	year = {2021},
	note = {arXiv:2110.14168 [cs.LG]},
}

@inproceedings{rodriguez_rankalign_2025,
	title = {{RankAlign}: {A} {Ranking} {View} of the {Generator}-{Validator} {Gap} in {Large} {Language} {Models}},
	shorttitle = {{RankAlign}},
	url = {https://openreview.net/forum?id=rJOkPauru9#discussion},
	language = {en},
	urldate = {2026-11-06},
	author = {Rodriguez, Juan Diego and Ding, Wenxuan and Erk, Katrin and Durrett, Greg},
	month = aug,
	year = {2025},
}

@misc{song_mind_2025,
	title = {Mind the {Gap}: {Examining} the {Self}-{Improvement} {Capabilities} of {Large} {Language} {Models}},
	shorttitle = {Mind the {Gap}},
	url = {http://arxiv.org/abs/2412.02674},
	doi = {10.48550/arXiv.2412.02674},
	urldate = {2026-11-04},
	publisher = {arXiv},
	author = {Song, Yuda and Zhang, Hanlin and Eisenach, Carson and Kakade, Sham and Foster, Dean and Ghai, Udaya},
	month = feb,
	year = {2025},
	note = {arXiv:2412.02674 [cs.CL]},
}

@misc{huang_large_2024,
	title = {Large {Language} {Models} {Cannot} {Self}-{Correct} {Reasoning} {Yet}},
	url = {http://arxiv.org/abs/2310.01798},
	doi = {10.48550/arXiv.2310.01798},
	urldate = {2026-11-04},
	publisher = {arXiv},
	author = {Huang, Jie and Chen, Xinyun and Mishra, Swaroop and Zheng, Huaixiu Steven and Yu, Adams Wei and Song, Xinying and Zhou, Denny},
	month = mar,
	year = {2024},
	note = {arXiv:2310.01798 [cs.CL]},
}

@inproceedings{madaan_selfrefine_2023,
	title = {Self-{Refine}: {Iterative} {Refinement} with {Self}-{Feedback}},
	volume = {36},
	shorttitle = {Self-{Refine}},
	url = {https://proceedings.neurips.cc/paper_files/paper/2023/hash/91edff07232fb1b55a505a9e9f6c0ff3-Abstract-Conference.html},
	doi = {10.52202/075280-2019},
	urldate = {2026-11-04},
	booktitle = {Advances in {Neural} {Information} {Processing} {Systems}},
	publisher = {Curran Associates, Inc.},
	author = {Madaan, Aman and Tandon, Niket and Gupta, Prakhar and Hallinan, Skyler and Gao, Luyu and Wiegreffe, Sarah and Alon, Uri and Dziri, Nouha and Prabhumoye, Shrimai and Yang, Yiming and Gupta, Shashank and Majumder, Bodhisattwa Prasad and Hermann, Katherine and Welleck, Sean and Yazdanbakhsh, Amir and Clark, Peter},
	year = {2023},
	pages = {46534--46594},
}

@article{xu_sensitivity_2007,
	title = {Sensitivity to sampling in {Bayesian} word learning},
	volume = {10},
	issn = {1467-7687},
	url = {https://onlinelibrary.wiley.com/doi/abs/10.1111/j.1467-7687.2007.00590.x},
	doi = {10.1111/j.1467-7687.2007.00590.x},
	language = {en},
	number = {3},
	urldate = {2026-10-21},
	journal = {Developmental Science},
	author = {Xu, Fei and Tenenbaum, Joshua B.},
	year = {2007},
	pages = {288--297},
}

@misc{bertolazzi_falsifybench_2026,
	title = {{FALSIFYBENCH}: {Evaluating} {Inductive} {Reasoning} in {LLMs} with {Rule} {Discovery} {Games}},
	shorttitle = {{FALSIFYBENCH}},
	url = {http://arxiv.org/abs/2606.04751},
	doi = {10.48550/arXiv.2606.04751},
	urldate = {2026-10-21},
	publisher = {arXiv},
	author = {Bertolazzi, Leonardo and Tentori, Katya and Bernardi, Raffaella},
	month = jun,
	year = {2026},
	note = {arXiv:2606.04751 [cs.AI]},
}

@misc{liu_hypobench_2026,
	title = {{HypoBench}: {Towards} {Systematic} and {Principled} {Benchmarking} for {Hypothesis} {Generation}},
	shorttitle = {{HypoBench}},
	url = {http://arxiv.org/abs/2504.11524},
	doi = {10.48550/arXiv.2504.11524},
	urldate = {2026-10-21},
	publisher = {arXiv},
	author = {Liu, Haokun and Huang, Sicong and Hu, Jingyu and Zhou, Yangqiaoyu and Tan, Chenhao},
	month = feb,
	year = {2026},
	note = {arXiv:2504.11524 [cs.AI]},
}

@misc{qiu_phenomenal_2024,
	title = {Phenomenal {Yet} {Puzzling}: {Testing} {Inductive} {Reasoning} {Capabilities} of {Language} {Models} with {Hypothesis} {Refinement}},
	shorttitle = {Phenomenal {Yet} {Puzzling}},
	url = {http://arxiv.org/abs/2310.08559},
	doi = {10.48550/arXiv.2310.08559},
	urldate = {2026-10-20},
	publisher = {arXiv},
	author = {Qiu, Linlu and Jiang, Liwei and Lu, Ximing and Sclar, Melanie and Pyatkin, Valentina and Bhagavatula, Chandra and Wang, Bailin and Kim, Yoon and Choi, Yejin and Dziri, Nouha and Ren, Xiang},
	month = may,
	year = {2024},
	note = {arXiv:2310.08559 [cs.CL]},
}

@misc{zweiger_selfadapting_2025,
	title = {Self-{Adapting} {Language} {Models}},
	url = {http://arxiv.org/abs/2506.10943},
	doi = {10.48550/arXiv.2506.10943},
	urldate = {2026-10-19},
	publisher = {arXiv},
	author = {Zweiger, Adam and Pari, Jyothish and Guo, Han and Akyürek, Ekin and Kim, Yoon and Agrawal, Pulkit},
	month = sep,
	year = {2025},
	note = {arXiv:2506.10943 [cs.LG]},
}

@article{gershman_amortized_2014,
	title = {Amortized {Inference} in {Probabilistic} {Reasoning}},
	volume = {36},
	url = {https://escholarship.org/uc/item/34j1h7k5},
	language = {en},
	number = {36},
	urldate = {2026-09-25},
	journal = {Proceedings of the Annual Meeting of the Cognitive Science Society},
	author = {Gershman, Samuel and Goodman, Noah},
	year = {2014},
}

@misc{murphy_model_2026,
	title = {Model {Discovery} {Agent}: {LLM}-assisted {Bayesian} experiment design for data-efficient discovery of mechanistic world models},
	shorttitle = {Model {Discovery} {Agent}},
	url = {http://arxiv.org/abs/2608.09696},
	doi = {10.48550/arXiv.2608.09696},
	urldate = {2026-09-09},
	publisher = {arXiv},
	author = {Murphy, Kevin},
	month = aug,
	year = {2026},
	note = {arXiv:2608.09696 [cs.AI]},
}

@misc{jagadish_closing_2026,
	title = {Closing the {Loop} to {Discover} {Psychological} {Theories} with an {Automated} {Cognitive} {Scientist}},
	url = {http://arxiv.org/abs/2606.26448},
	doi = {10.48550/arXiv.2606.26448},
	urldate = {2026-09-09},
	publisher = {arXiv},
	author = {Jagadish, Akshay K. and Strittmatter, Younes and Jacoby, Nori and Kachergis, George and Schulz, Eric and Daw, Nathaniel and Chandramouli, Suyog H. and Griffiths, Thomas L.},
	month = jun,
	year = {2026},
	note = {arXiv:2606.26448 [q-bio.NC]},
}

@inproceedings{wei_chainofthought_2022,
	title = {Chain-of-{Thought} {Prompting} {Elicits} {Reasoning} in {Large} {Language} {Models}},
	volume = {35},
	url = {https://proceedings.neurips.cc/paper/2022/hash/9d5609613524ecf4f15af0f7b31abca4-Abstract-Conference.html},
	doi = {10.52202/068431-1800},
	urldate = {2026-07-28},
	booktitle = {Advances in {Neural} {Information} {Processing} {Systems}},
	publisher = {Curran Associates, Inc.},
	author = {Wei, Jason and Wang, Xuezhi and Schuurmans, Dale and Bosma, Maarten and ichter, brian and Xia, Fei and Chi, Ed and Le, Quoc V and Zhou, Denny},
	year = {2022},
	pages = {24824--24837},
}

@misc{zhang_using_2026,
	title = {Using {Probabilistic} {Programs} to {Train} {Inductive} {Reasoning} in {Large} {Language} {Models}},
	url = {http://arxiv.org/abs/2606.09856},
	doi = {10.48550/arXiv.2606.09856},
	urldate = {2026-06-17},
	publisher = {arXiv},
	author = {Zhang, Liyi and Jagadish, Akshay K. and Lake, Brenden M. and Griffiths, Thomas L.},
	month = may,
	year = {2026},
	note = {arXiv:2606.09856 [cs.CL]},
}

@misc{kasenberg_aidiscovered_2026,
	title = {{AI}-{Discovered} {Cognitive} {Models} {Reveal} {Novel} {Insights} into {Human} and {Animal} {Learning}},
	copyright = {© 2026, Posted by openRxiv. This pre-print is available under a Creative Commons License (Attribution 4.0 International), CC BY 4.0, as described at http://creativecommons.org/licenses/by/4.0/},
	url = {https://www.biorxiv.org/content/10.64898/2026.05.18.725921v1},
	doi = {10.64898/2026.05.18.725921},
	language = {en},
	urldate = {2026-06-03},
	publisher = {bioRxiv},
	author = {Kasenberg, Daniel and Castro, Pablo Samuel and Eckstein, Maria K. and Éltető, Nóemi and Dabney, Will and Wang, Caroline and Engelcke, Martin and Mohanta, Rishika and Dev, Aparna and Botvinick, Matthew M. and Tomasev, Nenad and Turner, Glenn C. and Costa, Vincent and Daw, Nathaniel D. and Stachenfeld, Kimberly L. and Miller, Kevin J.},
	month = may,
	year = {2026},
	note = {ISSN: 2692-8205
Pages: 2026.05.18.725921
Section: New Results},
}

@misc{nvidia_nvidia_2025,
	title = {{NVIDIA} {Nemotron} 3: {Efficient} and {Open} {Intelligence}},
	shorttitle = {{NVIDIA} {Nemotron} 3},
	url = {http://arxiv.org/abs/2512.20856},
	doi = {10.48550/arXiv.2512.20856},
	urldate = {2026-05-07},
	publisher = {arXiv},
	author = {NVIDIA and Blakeman, Aaron and Grattafiori, Aaron and Basant, Aarti and Gupta, Abhibha and Khattar, Abhinav and Renduchintala, Adi and Vavre, Aditya and Shukla, Akanksha and Bercovich, Akhiad and Ficek, Aleksander and Shaposhnikov, Aleksandr and Kondratenko, Alex and Bukharin, Alexander and Milesi, Alexandre and Taghibakhshi, Ali and Liu, Alisa and Barton, Amelia and Mahabaleshwarkar, Ameya Sunil and Klein, Amir and Zuker, Amit and Geifman, Amnon and Shen, Amy and Bhiwandiwalla, Anahita and Tao, Andrew and Agrusa, Anjulie and Verma, Ankur and Guan, Ann and Mandarwal, Anubhav and Mehta, Arham and Aithal, Ashwath and Poojary, Ashwin and Ahamed, Asif and Mishra, Asit and Thekkumpate, Asma Kuriparambil and Dattagupta, Ayush and Zhu, Banghua and Sadeghi, Bardiya and Simkin, Barnaby and Lanir, Ben and Schifferer, Benedikt and Nushi, Besmira and Kartal, Bilal and Rouhani, Bita Darvish and Ginsburg, Boris and Norick, Brandon and Soubasis, Brandon and Kisacanin, Branislav and Yu, Brian and Catanzaro, Bryan and Mundo, Carlo del and Hwang, Chantal and Wang, Charles and Hsieh, Cheng-Ping and Zhang, Chenghao and Yu, Chenhan and Mungekar, Chetan and Patel, Chintan and Alexiuk, Chris and Parisien, Christopher and Neale, Collin and Meurillon, Cyril and Mosk-Aoyama, Damon and Su, Dan and Corneil, Dane and Afrimi, Daniel and Lo, Daniel and Rohrer, Daniel and Serebrenik, Daniel and Gitman, Daria and Levy, Daria and Stosic, Darko and Mosallanezhad, David and Narayanan, Deepak and Nathawani, Dhruv and Rekesh, Dima and Yared, Dina and Kakwani, Divyanshu and Ahn, Dong and Riach, Duncan and Stosic, Dusan and Minasyan, Edgar and Lin, Edward and Long, Eileen and Long, Eileen Peters and Segal, Elad and Lantz, Elena and Evans, Ellie and Ning, Elliott and Chung, Eric and Harper, Eric and Tramel, Eric and Galinkin, Erick and Pounds, Erik and Briones, Evan and Bakhturina, Evelina and Tsykunov, Evgeny and Ladhak, Faisal and Wang, Fay and Jia, Fei and Soares, Felipe and Chen, Feng and Galko, Ferenc and Sun, Frank and Siino, Frankie and Agam, Gal Hubara and Ajjanagadde, Ganesh and Bhatt, Gantavya and Prasad, Gargi and Armstrong, George and Shen, Gerald and Batmaz, Gorkem and Nalbandyan, Grigor and Qian, Haifeng and Sharma, Harsh and Ross, Hayley and Ngo, Helen and Hum, Herbert and Sahota, Herman and Wang, Hexin and Soni, Himanshu and Upadhyay, Hiren and Mao, Huizi and Nguyen, Huy C. and Nguyen, Huy Q. and Cunningham, Iain and Galil, Ido and Shahaf, Ido and Gitman, Igor and Loshchilov, Ilya and Schen, Itamar and Levy, Itay and Moshkov, Ivan and Golan, Izik and Putterman, Izzy and Kautz, Jan and Scowcroft, Jane Polak and Casper, Jared and Mitra, Jatin and Glick, Jeffrey and Chen, Jenny and Oliver, Jesse and Zhang, Jian and Zeng, Jiaqi and Lou, Jie and Zhang, Jimmy and Choi, Jinhang and Huang, Jining and Conway, Joey and Guman, Joey and Kamalu, John and Greco, Johnny and Cohen, Jonathan and Jennings, Joseph and Daw, Joyjit and Vialard, Julien Veron and Yi, Junkeun and Parmar, Jupinder and Xu, Kai and Zhu, Kan and Briski, Kari and Cheung, Katherine and Luna, Katherine and Wyss, Keith and Santhanam, Keshav and Shih, Kevin and Kong, Kezhi and Bhardwaj, Khushi and Shankar, Kirthi and Puvvada, Krishna C. and Pawelec, Krzysztof and Anik, Kumar and McAfee, Lawrence and Sleiman, Laya and Derczynski, Leon and Ding, Li and Wei, Lizzie and Liebenwein, Lucas and Vega, Luis and Grover, Maanu and Segbroeck, Maarten Van and Melo, Maer Rodrigues de and Nazemi, Mahdi and Sreedhar, Makesh Narsimhan and Kilaru, Manoj and Ashkenazi, Maor and Romeijn, Marc and Chochowski, Marcin and Cai, Mark and Kliegl, Markus and Moosaei, Maryam and Kulka, Matt and Novikov, Matvei and Samadi, Mehrzad and Corpuz, Melissa and Wang, Mengru and Price, Meredith and Andersch, Michael and Boone, Michael and Evans, Michael and Martinez, Miguel and Khona, Mikail and Chrzanowski, Mike and Lee, Minseok and Dabbah, Mohammad and Shoeybi, Mohammad and Patwary, Mostofa and Mulepati, Nabin and Nabwani, Najeeb and Hereth, Natalie and Assaf, Nave and Habibi, Negar and Zmora, Neta and Haber, Netanel and Sessions, Nicola and Bhatia, Nidhi and Jukar, Nikhil and Pope, Nikki and Ludwig, Nikolai and Tajbakhsh, Nima and Ailon, Nir and Juluru, Nirmal and Sharma, Nishant and Hrinchuk, Oleksii and Kuchaiev, Oleksii and Delalleau, Olivier and Olabiyi, Oluwatobi and Argov, Omer Ullman and Puny, Omri and Tropp, Oren and Xie, Ouye and Chadha, Parth and Shamis, Pasha and Gibbons, Paul and Molchanov, Pavlo and Morkisz, Pawel and Dykas, Peter and Jin, Peter and Xu, Pinky and Januszewski, Piotr and Thombre, Pranav Prashant and Varshney, Prasoon and Gundecha, Pritam and Tredak, Przemek and Miao, Qing and Wan, Qiyu and Mahabadi, Rabeeh Karimi and Garg, Rachit and El-Yaniv, Ran and Zilberstein, Ran and Shafipour, Rasoul and Harang, Rich and Izzo, Rick and Shahbazyan, Rima and Garg, Rishabh and Borkar, Ritika and Gala, Ritu and Islam, Riyad and Hesse, Robert and Waleffe, Roger and Watve, Rohit and Koren, Roi and Zhang, Ruoxi and Hewett, Russell and Hewett, Russell J. and Prenger, Ryan and Timbrook, Ryan and Mahdavi, Sadegh and Modi, Sahil and Kriman, Samuel and Lim, Sangkug and Kariyappa, Sanjay and Satheesh, Sanjeev and Kaji, Saori and Pasumarthi, Satish and Muralidharan, Saurav and Narentharen, Sean and Narenthiran, Sean and Bak, Seonmyeong and Kashirsky, Sergey and Poulos, Seth and Mor, Shahar and Ramasamy, Shanmugam and Acharya, Shantanu and Ghosh, Shaona and Sreenivas, Sharath Turuvekere and Thomas, Shelby and Fan, Shiqing and Gopal, Shreya and Prabhumoye, Shrimai and Pachori, Shubham and Toshniwal, Shubham and Ding, Shuoyang and Singh, Siddharth and Sun, Simeng and Ithape, Smita and Majumdar, Somshubra and Singhal, Soumye and Sergienko, Stas and Alborghetti, Stefania and Ge, Stephen and Devare, Sugam Dipak and Barua, Sumeet Kumar and Panguluri, Suseella and Gupta, Suyog and Priyadarshi, Sweta and Akter, Syeda Nahida and Bui, Tan and Ene, Teodor-Dumitru and Kong, Terry and Do, Thanh and Blankevoort, Tijmen and Moon, Tim and Balough, Tom and Asida, Tomer and Natan, Tomer Bar and Ronen, Tomer and Konuk, Tugrul and Vashishth, Twinkle and Karpas, Udi and De, Ushnish and Noorozi, Vahid and Noroozi, Vahid and Srinivasan, Venkat and Elango, Venmugil and Cui, Victor and Korthikanti, Vijay and Rao, Vinay and Kurin, Vitaly and Lavrukhin, Vitaly and Anisimov, Vladimir and Jiang, Wanli and Ahmad, Wasi Uddin and Du, Wei and Ping, Wei and Zhou, Wenfei and Jennings, Will and Zhang, William and Prazuch, Wojciech and Ren, Xiaowei and Karnati, Yashaswi and Choi, Yejin and Meyer, Yev and Wu, Yi-Fu and Zhang, Yian and Qin, Yigong and Lin, Ying and Geifman, Yonatan and Fu, Yonggan and Subara, Yoshi and Suhara, Yoshi and Gao, Yubo and Moshe, Zach and Dong, Zhen and Zhu, Zhongbo and Liu, Zihan and Chen, Zijia and Yan, Zijie},
	month = dec,
	year = {2025},
	note = {arXiv:2512.20856 [cs]},
}

@misc{gemma_2026,
	title = {Gemma 4: {Byte} for byte, the most capable open models},
	shorttitle = {Gemma 4},
	url = {https://blog.google/innovation-and-ai/technology/developers-tools/gemma-4/},
	language = {en-us},
	urldate = {2026-05-07},
	journal = {Google},
	author = {Gemma Team},
	month = apr,
	year = {2026},
}

@misc{mitchener_kosmos_2025,
	title = {Kosmos: {An} {AI} {Scientist} for {Autonomous} {Discovery}},
	shorttitle = {Kosmos},
	url = {http://arxiv.org/abs/2511.02824},
	doi = {10.48550/arXiv.2511.02824},
	urldate = {2026-05-04},
	publisher = {arXiv},
	author = {Mitchener, Ludovico and Yiu, Angela and Chang, Benjamin and Bourdenx, Mathieu and Nadolski, Tyler and Sulovari, Arvis and Landsness, Eric C. and Barabasi, Daniel L. and Narayanan, Siddharth and Evans, Nicky and Reddy, Shriya and Foiani, Martha and Kamal, Aizad and Shriver, Leah P. and Cao, Fang and Wassie, Asmamaw T. and Laurent, Jon M. and Melville-Green, Edwin and Caldas, Mayk and Bou, Albert and Roberts, Kaleigh F. and Zagorac, Sladjana and Orr, Timothy C. and Orr, Miranda E. and Zwezdaryk, Kevin J. and Ghareeb, Ali E. and McCoy, Laurie and Gomes, Bruna and Ashley, Euan A. and Duff, Karen E. and Buonassisi, Tonio and Rainforth, Tom and Bateman, Randall J. and Skarlinski, Michael and Rodriques, Samuel G. and Hinks, Michaela M. and White, Andrew D.},
	month = nov,
	year = {2025},
	note = {arXiv:2511.02824 [cs]},
}

@misc{yamada_ai_2025,
	title = {The {AI} {Scientist}-v2: {Workshop}-{Level} {Automated} {Scientific} {Discovery} via {Agentic} {Tree} {Search}},
	shorttitle = {The {AI} {Scientist}-v2},
	url = {http://arxiv.org/abs/2504.08066},
	doi = {10.48550/arXiv.2504.08066},
	urldate = {2026-05-04},
	publisher = {arXiv},
	author = {Yamada, Yutaro and Lange, Robert Tjarko and Lu, Cong and Hu, Shengran and Lu, Chris and Foerster, Jakob and Clune, Jeff and Ha, David},
	month = apr,
	year = {2025},
	note = {arXiv:2504.08066 [cs]},
}

@incollection{clark_whats_1973,
	title = {What's in a word? {On} the child's acquisition of semantics in his first language},
	shorttitle = {What's in a word?},
	url = {https://www.sciencedirect.com/science/article/pii/B9780125058506500098},
	urldate = {2026-05-04},
	booktitle = {Cognitive development and acquisition of language},
	publisher = {Elsevier},
	author = {Clark, Eve V.},
	year = {1973},
	pages = {65--110},
}

@article{macnamara_cognitive_1972,
	address = {US},
	title = {Cognitive basis of language learning in infants},
	volume = {79},
	issn = {1939-1471},
	doi = {10.1037/h0031901},
	number = {1},
	journal = {Psychological Review},
	publisher = {American Psychological Association},
	author = {MacNamara, John},
	year = {1972},
	pages = {1--13},
}

@article{rescorla_overextension_1980,
	title = {Overextension in early language development},
	volume = {7},
	issn = {1469-7602, 0305-0009},
	url = {https://www.cambridge.org/core/journals/journal-of-child-language/article/abs/overextension-in-early-language-development/B42AE23A10088ACCBA90AE2E88FD91A7},
	doi = {10.1017/S0305000900002658},
	language = {en},
	number = {2},
	urldate = {2026-05-04},
	journal = {Journal of Child Language},
	author = {Rescorla, Leslie A.},
	month = jun,
	year = {1980},
	pages = {321--335},
}

@incollection{vonhelmholtz_facts_1878,
	address = {Connecticut},
	title = {The {Facts} of {Perception}},
	isbn = {978-0-8195-4039-3},
	language = {en},
	urldate = {2026-05-03},
	booktitle = {Selected {Writings} of {Hermann} von {Helmholtz}},
	publisher = {Wesleyan University Press},
	author = {von Helmholtz, Hermann},
	editor = {Russell, Kahl},
	year = {1878},
}

@inproceedings{bigelow_inferring_2016,
	title = {Inferring priors in compositional cognitive models},
	volume = {38},
	url = {https://escholarship.org/uc/item/43s5z8jj},
	language = {en},
	number = {0},
	urldate = {2026-04-07},
	booktitle = {Proceedings of the {Annual} {Meeting} of the {Cognitive} {Science} {Society}},
	author = {Bigelow, Eric J. and Piantadosi, Steven T.},
	year = {2016},
}

@article{bigelow_large_2016,
	title = {A large dataset of generalization patterns in the number game},
	volume = {4},
	url = {https://account.openpsychologydata.metajnl.com/index.php/up-j-jopd/article/view/jopd.19},
	doi = {10.5334/jopd.19},
	number = {1},
	urldate = {2026-04-06},
	journal = {Journal of Open Psychology Data},
	author = {Bigelow, Eric and Piantadosi, Steven T.},
	year = {2016},
	pages = {e4--e4},
}

@inproceedings{tenenbaum_rules_1999,
	title = {Rules and {Similarity} in {Concept} {Learning}},
	volume = {12},
	url = {https://proceedings.neurips.cc/paper/1999/hash/86d7c8a08b4aaa1bc7c599473f5dddda-Abstract.html},
	urldate = {2026-04-06},
	booktitle = {Advances in {Neural} {Information} {Processing} {Systems}},
	publisher = {MIT Press},
	author = {Tenenbaum, Joshua},
	year = {1999},
}

@article{qiu_bayesian_2026,
	title = {Bayesian teaching enables probabilistic reasoning in large language models},
	volume = {17},
	copyright = {2026 The Author(s)},
	issn = {2041-1723},
	url = {https://www.nature.com/articles/s41467-025-67998-6},
	doi = {10.1038/s41467-025-67998-6},
	language = {en},
	number = {1},
	urldate = {2026-04-06},
	journal = {Nature Communications},
	publisher = {Nature Publishing Group},
	author = {Qiu, Linlu and Sha, Fei and Allen, Kelsey and Kim, Yoon and Linzen, Tal and van Steenkiste, Sjoerd},
	month = jan,
	year = {2026},
	pages = {1238},
}

@misc{padmanabhan_language_2025,
	title = {On {Language} {Models}' {Sensitivity} to {Suspicious} {Coincidences}},
	url = {http://arxiv.org/abs/2504.09387},
	doi = {10.48550/arXiv.2504.09387},
	urldate = {2026-04-06},
	publisher = {arXiv},
	author = {Padmanabhan, Sriram and Misra, Kanishka and Mahowald, Kyle and Choi, Eunsol},
	month = apr,
	year = {2025},
	note = {arXiv:2504.09387 [cs]},
}

@misc{bazigaran_concept_2025,
	title = {Concept {Generalization} in {Humans} and {Large} {Language} {Models}: {Insights} from the {Number} {Game}},
	shorttitle = {Concept {Generalization} in {Humans} and {Large} {Language} {Models}},
	url = {http://arxiv.org/abs/2512.20162},
	doi = {10.48550/arXiv.2512.20162},
	urldate = {2026-04-05},
	publisher = {arXiv},
	author = {Bazigaran, Arghavan and Sohn, Hansem},
	month = dec,
	year = {2025},
	note = {arXiv:2512.20162 [cs]},
}

@misc{pournemat_reasoning_2025,
	title = {Reasoning {Under} {Uncertainty}: {Exploring} {Probabilistic} {Reasoning} {Capabilities} of {LLMs}},
	shorttitle = {Reasoning {Under} {Uncertainty}},
	url = {http://arxiv.org/abs/2509.10739},
	doi = {10.48550/arXiv.2509.10739},
	urldate = {2026-04-05},
	publisher = {arXiv},
	author = {Pournemat, Mobina and Rezaei, Keivan and Sriramanan, Gaurang and Zarei, Arman and Fu, Jiaxiang and Wang, Yang and Eghbalzadeh, Hamid and Feizi, Soheil},
	month = sep,
	year = {2025},
	note = {arXiv:2509.10739 [cs]},
}

@misc{liu_cognitive_2026,
	title = {Cognitive {Models} and {AI} {Algorithms} {Provide} {Templates} for {Designing} {Language} {Agents}},
	url = {http://arxiv.org/abs/2602.22523},
	doi = {10.48550/arXiv.2602.22523},
	urldate = {2026-04-04},
	publisher = {arXiv},
	author = {Liu, Ryan and Arumugam, Dilip and Zhang, Cedegao E. and Escola, Sean and Pitkow, Xaq and Griffiths, Thomas L.},
	month = feb,
	year = {2026},
	note = {arXiv:2602.22523 [cs]},
}

@phdthesis{tenenbaum_bayesian_1999,
	type = {Thesis},
	title = {A {Bayesian} framework for concept learning},
	copyright = {M.I.T. theses are protected by copyright. They may be viewed from this source for any purpose, but reproduction or distribution in any format is prohibited without written permission. See provided URL for inquiries about permission.},
	issn = {4247-1842},
	url = {https://dspace.mit.edu/handle/1721.1/16714},
	language = {eng},
	urldate = {2026-03-29},
	school = {Massachusetts Institute of Technology},
	author = {Tenenbaum, Joshua B. (Joshua Brett)},
	year = {1999},
	note = {Accepted: 2005-05-19T14:18:52Z},
}

@misc{qwen_qwen35_2026,
	title = {Qwen3.5: {Towards} {Native} {Multimodal} {Agents}},
	copyright = {Apache-2.0},
	url = {https://qwen.ai/blog?id=qwen3.5},
	urldate = {2026-03-20},
	author = {Qwen, Team},
	month = feb,
	year = {2026},
	note = {original-date: 2025-09-11T05:32:39Z},
}

@article{xu_word_2007,
	title = {Word learning as {Bayesian} inference.},
	volume = {114},
	issn = {1939-1471, 0033-295X},
	url = {http://doi.apa.org/getdoi.cfm?doi=10.1037/0033-295X.114.2.245},
	doi = {10.1037/0033-295X.114.2.245},
	language = {en},
	number = {2},
	urldate = {2022-11-07},
	journal = {Psychological Review},
	author = {Xu, Fei and Tenenbaum, Joshua B.},
	year = {2007},
	pages = {245--272},
}

@article{tenenbaum_how_2011,
	title = {How to {Grow} a {Mind}: {Statistics}, {Structure}, and {Abstraction}},
	volume = {331},
	issn = {0036-8075, 1095-9203},
	shorttitle = {How to {Grow} a {Mind}},
	url = {https://www.science.org/doi/10.1126/science.1192788},
	doi = {10.1126/science.1192788},
	language = {en},
	number = {6022},
	urldate = {2022-11-07},
	journal = {Science},
	author = {Tenenbaum, Joshua B. and Kemp, Charles and Griffiths, Thomas L. and Goodman, Noah D.},
	month = mar,
	year = {2011},
	pages = {1279--1285},
}

@article{lake_humanlevel_2015,
	title = {Human-level concept learning through probabilistic program induction},
	volume = {350},
	issn = {0036-8075, 1095-9203},
	url = {https://www.sciencemag.org/lookup/doi/10.1126/science.aab3050},
	doi = {10.1126/science.aab3050},
	language = {en},
	number = {6266},
	urldate = {2022-11-07},
	journal = {Science},
	author = {Lake, B. M. and Salakhutdinov, R. and Tenenbaum, J. B.},
	month = dec,
	year = {2015},
	pages = {1332--1338},
}

@article{dasgupta_memory_2021,
	title = {Memory as a {Computational} {Resource}},
	volume = {25},
	issn = {13646613},
	url = {https://linkinghub.elsevier.com/retrieve/pii/S1364661320303053},
	doi = {10.1016/j.tics.2020.12.008},
	language = {en},
	number = {3},
	urldate = {2022-11-07},
	journal = {Trends in Cognitive Sciences},
	author = {Dasgupta, Ishita and Gershman, Samuel J.},
	month = mar,
	year = {2021},
	pages = {240--251},
}

@misc{xie_explanation_2022,
	title = {An {Explanation} of {In}-context {Learning} as {Implicit} {Bayesian} {Inference}},
	url = {http://arxiv.org/abs/2111.02080},
	doi = {10.48550/arXiv.2111.02080},
	urldate = {2023-10-24},
	publisher = {arXiv},
	author = {Xie, Sang Michael and Raghunathan, Aditi and Liang, Percy and Ma, Tengyu},
	month = jul,
	year = {2022},
	note = {arXiv:2111.02080 [cs]},
}

@misc{ellis_humanlike_2023,
	title = {Human-like {Few}-{Shot} {Learning} via {Bayesian} {Reasoning} over {Natural} {Language}},
	url = {http://arxiv.org/abs/2306.02797},
	doi = {10.48550/arXiv.2306.02797},
	urldate = {2023-12-19},
	publisher = {arXiv},
	author = {Ellis, Kevin},
	month = sep,
	year = {2023},
	note = {arXiv:2306.02797 [cs]},
}

@misc{muller_transformers_2021,
	title = {Transformers {Can} {Do} {Bayesian} {Inference}},
	url = {https://arxiv.org/abs/2112.10510v6},
	language = {en},
	urldate = {2024-01-02},
	author = {Müller, Samuel and Hollmann, Noah and Arango, Sebastian Pineda and Grabocka, Josif and Hutter, Frank},
	month = dec,
	year = {2021},
}

@article{lake_humanlike_2023,
	title = {Human-like systematic generalization through a meta-learning neural network},
	volume = {623},
	copyright = {2023 The Author(s)},
	issn = {1476-4687},
	url = {https://www.nature.com/articles/s41586-023-06668-3},
	doi = {10.1038/s41586-023-06668-3},
	language = {en},
	number = {7985},
	urldate = {2024-01-07},
	journal = {Nature},
	publisher = {Nature Publishing Group},
	author = {Lake, Brenden M. and Baroni, Marco},
	month = nov,
	year = {2023},
	note = {Number: 7985},
	pages = {115--121},
}

@inproceedings{lake_generalization_2018,
	title = {Generalization without {Systematicity}: {On} the {Compositional} {Skills} of {Sequence}-to-{Sequence} {Recurrent} {Networks}},
	issn = {2640-3498},
	shorttitle = {Generalization without {Systematicity}},
	url = {https://proceedings.mlr.press/v80/lake18a.html},
	language = {en},
	urldate = {2024-01-09},
	booktitle = {Proceedings of the 35th {International} {Conference} on {Machine} {Learning}},
	publisher = {PMLR},
	author = {Lake, Brenden and Baroni, Marco},
	month = jul,
	year = {2018},
	pages = {2873--2882},
}

@misc{gandhi_boxinggym_2025,
	title = {{BoxingGym}: {Benchmarking} {Progress} in {Automated} {Experimental} {Design} and {Model} {Discovery}},
	shorttitle = {{BoxingGym}},
	url = {http://arxiv.org/abs/2501.01540},
	doi = {10.48550/arXiv.2501.01540},
	urldate = {2025-01-09},
	publisher = {arXiv},
	author = {Gandhi, Kanishk and Li, Michael Y. and Goodyear, Lyle and Li, Louise and Bhaskar, Aditi and Zaman, Mohammed and Goodman, Noah D.},
	month = jan,
	year = {2025},
	note = {arXiv:2501.01540 [cs]},
}

@article{navarro_sampling_2012,
	title = {Sampling {Assumptions} in {Inductive} {Generalization}},
	volume = {36},
	copyright = {Copyright © 2011 Cognitive Science Society, Inc.},
	issn = {1551-6709},
	url = {https://onlinelibrary.wiley.com/doi/abs/10.1111/j.1551-6709.2011.01212.x},
	doi = {10.1111/j.1551-6709.2011.01212.x},
	language = {en},
	number = {2},
	urldate = {2025-06-01},
	journal = {Cognitive Science},
	author = {Navarro, Daniel J. and Dry, Matthew J. and Lee, Michael D.},
	year = {2012},
	note = {\_eprint: https://onlinelibrary.wiley.com/doi/pdf/10.1111/j.1551-6709.2011.01212.x},
	pages = {187--223},
}

@misc{battleday_artificial_2024,
	title = {Artificial intelligence for science: {The} easy and hard problems},
	shorttitle = {Artificial intelligence for science},
	url = {http://arxiv.org/abs/2408.14508},
	doi = {10.48550/arXiv.2408.14508},
	urldate = {2025-07-17},
	publisher = {arXiv},
	author = {Battleday, Ruairidh M. and Gershman, Samuel J.},
	month = dec,
	year = {2024},
	note = {arXiv:2408.14508 [cs]},
}

@article{cornelio_combining_2023,
	title = {Combining data and theory for derivable scientific discovery with {AI}-{Descartes}},
	volume = {14},
	copyright = {2023 The Author(s)},
	issn = {2041-1723},
	url = {https://www.nature.com/articles/s41467-023-37236-y},
	doi = {10.1038/s41467-023-37236-y},
	language = {en},
	number = {1},
	urldate = {2025-07-17},
	journal = {Nature Communications},
	publisher = {Nature Publishing Group},
	author = {Cornelio, Cristina and Dash, Sanjeeb and Austel, Vernon and Josephson, Tyler R. and Goncalves, Joao and Clarkson, Kenneth L. and Megiddo, Nimrod and El Khadir, Bachir and Horesh, Lior},
	month = apr,
	year = {2023},
	pages = {1777},
}

@article{guo_deepseekr1_2025,
	title = {{DeepSeek}-{R1} incentivizes reasoning in {LLMs} through reinforcement learning},
	volume = {645},
	copyright = {2025 The Author(s)},
	issn = {1476-4687},
	url = {https://www.nature.com/articles/s41586-025-09422-z},
	doi = {10.1038/s41586-025-09422-z},
	language = {en},
	number = {8081},
	urldate = {2025-09-23},
	journal = {Nature},
	publisher = {Nature Publishing Group},
	author = {Guo, Daya and Yang, Dejian and Zhang, Haowei and Song, Junxiao and Wang, Peiyi and Zhu, Qihao and Xu, Runxin and Zhang, Ruoyu and Ma, Shirong and Bi, Xiao and Zhang, Xiaokang and Yu, Xingkai and Wu, Yu and Wu, Z. F. and Gou, Zhibin and Shao, Zhihong and Li, Zhuoshu and Gao, Ziyi and Liu, Aixin and Xue, Bing and Wang, Bingxuan and Wu, Bochao and Feng, Bei and Lu, Chengda and Zhao, Chenggang and Deng, Chengqi and Ruan, Chong and Dai, Damai and Chen, Deli and Ji, Dongjie and Li, Erhang and Lin, Fangyun and Dai, Fucong and Luo, Fuli and Hao, Guangbo and Chen, Guanting and Li, Guowei and Zhang, H. and Xu, Hanwei and Ding, Honghui and Gao, Huazuo and Qu, Hui and Li, Hui and Guo, Jianzhong and Li, Jiashi and Chen, Jingchang and Yuan, Jingyang and Tu, Jinhao and Qiu, Junjie and Li, Junlong and Cai, J. L. and Ni, Jiaqi and Liang, Jian and Chen, Jin and Dong, Kai and Hu, Kai and You, Kaichao and Gao, Kaige and Guan, Kang and Huang, Kexin and Yu, Kuai and Wang, Lean and Zhang, Lecong and Zhao, Liang and Wang, Litong and Zhang, Liyue and Xu, Lei and Xia, Leyi and Zhang, Mingchuan and Zhang, Minghua and Tang, Minghui and Zhou, Mingxu and Li, Meng and Wang, Miaojun and Li, Mingming and Tian, Ning and Huang, Panpan and Zhang, Peng and Wang, Qiancheng and Chen, Qinyu and Du, Qiushi and Ge, Ruiqi and Zhang, Ruisong and Pan, Ruizhe and Wang, Runji and Chen, R. J. and Jin, R. L. and Chen, Ruyi and Lu, Shanghao and Zhou, Shangyan and Chen, Shanhuang and Ye, Shengfeng and Wang, Shiyu and Yu, Shuiping and Zhou, Shunfeng and Pan, Shuting and Li, S. S. and Zhou, Shuang and Wu, Shaoqing and Yun, Tao and Pei, Tian and Sun, Tianyu and Wang, T. and Zeng, Wangding and Liu, Wen and Liang, Wenfeng and Gao, Wenjun and Yu, Wenqin and Zhang, Wentao and Xiao, W. L. and An, Wei and Liu, Xiaodong and Wang, Xiaohan and Chen, Xiaokang and Nie, Xiaotao and Cheng, Xin and Liu, Xin and Xie, Xin and Liu, Xingchao and Yang, Xinyu and Li, Xinyuan and Su, Xuecheng and Lin, Xuheng and Li, X. Q. and Jin, Xiangyue and Shen, Xiaojin and Chen, Xiaosha and Sun, Xiaowen and Wang, Xiaoxiang and Song, Xinnan and Zhou, Xinyi and Wang, Xianzu and Shan, Xinxia and Li, Y. K. and Wang, Y. Q. and Wei, Y. X. and Zhang, Yang and Xu, Yanhong and Li, Yao and Zhao, Yao and Sun, Yaofeng and Wang, Yaohui and Yu, Yi and Zhang, Yichao and Shi, Yifan and Xiong, Yiliang and He, Ying and Piao, Yishi and Wang, Yisong and Tan, Yixuan and Ma, Yiyang and Liu, Yiyuan and Guo, Yongqiang and Ou, Yuan and Wang, Yuduan and Gong, Yue and Zou, Yuheng and He, Yujia and Xiong, Yunfan and Luo, Yuxiang and You, Yuxiang and Liu, Yuxuan and Zhou, Yuyang and Zhu, Y. X. and Huang, Yanping and Li, Yaohui and Zheng, Yi and Zhu, Yuchen and Ma, Yunxian and Tang, Ying and Zha, Yukun and Yan, Yuting and Ren, Z. Z. and Ren, Zehui and Sha, Zhangli and Fu, Zhe and Xu, Zhean and Xie, Zhenda and Zhang, Zhengyan and Hao, Zhewen and Ma, Zhicheng and Yan, Zhigang and Wu, Zhiyu and Gu, Zihui and Zhu, Zijia and Liu, Zijun and Li, Zilin and Xie, Ziwei and Song, Ziyang and Pan, Zizheng and Huang, Zhen and Xu, Zhipeng and Zhang, Zhongyu and Zhang, Zhen},
	month = sep,
	year = {2025},
	pages = {633--638},
}

@article{fodor_connectionism_1988,
	title = {Connectionism and cognitive architecture: {A} critical analysis},
	volume = {28},
	issn = {0010-0277},
	shorttitle = {Connectionism and cognitive architecture},
	url = {https://www.sciencedirect.com/science/article/pii/0010027788900315},
	doi = {10.1016/0010-0277(88)90031-5},
	number = {1},
	urldate = {2025-09-24},
	journal = {Cognition},
	author = {Fodor, Jerry A. and Pylyshyn, Zenon W.},
	month = mar,
	year = {1988},
	pages = {3--71},
}

@misc{novikov_alphaevolve_2025,
	title = {{AlphaEvolve}: {A} coding agent for scientific and algorithmic discovery},
	shorttitle = {{AlphaEvolve}},
	url = {http://arxiv.org/abs/2506.13131},
	doi = {10.48550/arXiv.2506.13131},
	urldate = {2025-10-12},
	publisher = {arXiv},
	author = {Novikov, Alexander and Vũ, Ngân and Eisenberger, Marvin and Dupont, Emilien and Huang, Po-Sen and Wagner, Adam Zsolt and Shirobokov, Sergey and Kozlovskii, Borislav and Ruiz, Francisco J. R. and Mehrabian, Abbas and Kumar, M. Pawan and See, Abigail and Chaudhuri, Swarat and Holland, George and Davies, Alex and Nowozin, Sebastian and Kohli, Pushmeet and Balog, Matej},
	month = jun,
	year = {2025},
	note = {arXiv:2506.13131 [cs]},
}

@misc{gottweis_ai_2025,
	title = {Towards an {AI} co-scientist},
	url = {http://arxiv.org/abs/2502.18864},
	doi = {10.48550/arXiv.2502.18864},
	urldate = {2025-10-21},
	publisher = {arXiv},
	author = {Gottweis, Juraj and Weng, Wei-Hung and Daryin, Alexander and Tu, Tao and Palepu, Anil and Sirkovic, Petar and Myaskovsky, Artiom and Weissenberger, Felix and Rong, Keran and Tanno, Ryutaro and Saab, Khaled and Popovici, Dan and Blum, Jacob and Zhang, Fan and Chou, Katherine and Hassidim, Avinatan and Gokturk, Burak and Vahdat, Amin and Kohli, Pushmeet and Matias, Yossi and Carroll, Andrew and Kulkarni, Kavita and Tomasev, Nenad and Guan, Yuan and Dhillon, Vikram and Vaishnav, Eeshit Dhaval and Lee, Byron and Costa, Tiago R. D. and Penadés, José R. and Peltz, Gary and Xu, Yunhan and Pawlosky, Annalisa and Karthikesalingam, Alan and Natarajan, Vivek},
	month = feb,
	year = {2025},
	note = {arXiv:2502.18864 [cs]},
}

@inproceedings{brown_language_2020,
	title = {Language {Models} are {Few}-{Shot} {Learners}},
	volume = {33},
	url = {https://proceedings.neurips.cc/paper_files/paper/2020/hash/1457c0d6bfcb4967418bfb8ac142f64a-Abstract.html?utm_source=transaction&utm_medium=email&utm_campaign=linkedin_newsletter},
	language = {en},
	urldate = {2025-10-23},
	booktitle = {Advances in {Neural} {Information} {Processing} {Systems}},
	author = {Brown, Tom and Mann, Benjamin and Ryder, Nick and Subbiah, Melanie and Kaplan, Jared D. and Dhariwal, Prafulla and Neelakantan, Arvind and Shyam, Pranav and Sastry, Girish and Askell, Amanda and Agarwal, Sandhini and Herbert-Voss, Ariel and Krueger, Gretchen and Henighan, Tom and Child, Rewon and Ramesh, Aditya and Ziegler, Daniel and Wu, Jeffrey and Winter, Clemens and Hesse, Chris and Chen, Mark and Sigler, Eric and Litwin, Mateusz and Gray, Scott and Chess, Benjamin and Clark, Jack and Berner, Christopher and McCandlish, Sam and Radford, Alec and Sutskever, Ilya and Amodei, Dario},
	year = {2020},
	pages = {1877--1901},
}

@book{popper_logic_2005,
	title = {The logic of scientific discovery},
	url = {https://www.taylorfrancis.com/books/mono/10.4324/9780203994627/logic-scientific-discovery-karl-popper-karl-popper},
	urldate = {2025-10-25},
	publisher = {Routledge},
	author = {Popper, Karl},
	year = {2005},
}

@incollection{lakatos_falsification_2014,
	title = {Falsification and the methodology of scientific research programmes},
	url = {https://api.taylorfrancis.com/content/chapters/edit/download?identifierName=doi&identifierValue=10.4324/9780203802458-7&type=chapterpdf},
	urldate = {2025-10-25},
	booktitle = {Philosophy, science, and history},
	publisher = {Routledge},
	author = {Lakatos, Imre},
	year = {2014},
	pages = {89--94},
}

@article{oh_discovering_2025,
	title = {Discovering state-of-the-art reinforcement learning algorithms},
	copyright = {2025 The Author(s), under exclusive licence to Springer Nature Limited},
	issn = {1476-4687},
	url = {https://www.nature.com/articles/s41586-025-09761-x},
	doi = {10.1038/s41586-025-09761-x},
	language = {en},
	urldate = {2025-11-06},
	journal = {Nature},
	publisher = {Nature Publishing Group},
	author = {Oh, Junhyuk and Farquhar, Greg and Kemaev, Iurii and Calian, Dan A. and Hessel, Matteo and Zintgraf, Luisa and Singh, Satinder and van Hasselt, Hado and Silver, David},
	month = oct,
	year = {2025},
	pages = {1--2},
}

@misc{zhang_darwin_2026,
	title = {Darwin {Godel} {Machine}: {Open}-{Ended} {Evolution} of {Self}-{Improving} {Agents}},
	shorttitle = {Darwin {Godel} {Machine}},
	url = {http://arxiv.org/abs/2505.22954},
	doi = {10.48550/arXiv.2505.22954},
	urldate = {2026-03-16},
	publisher = {arXiv},
	author = {Zhang, Jenny and Hu, Shengran and Lu, Cong and Lange, Robert and Clune, Jeff},
	month = mar,
	year = {2026},
	note = {arXiv:2505.22954 [cs]},
}


\clearpage

\appendix

\section{Models and prompt conditions}
\label{sec:appendix_models_prompts}

The model names in the main text are short aliases for the following identifiers: Gemma 4 A4B, \texttt{google/gemma-4-26B-A4B-it}, and Gemma 4 E4B, \texttt{google/gemma-4-E4B-it} \citep{gemma_2026}; Qwen 3.6 A3B, \texttt{Qwen/Qwen3.6-35B-A3B}, and Qwen 3.5 4B, \texttt{Qwen/Qwen3.5-4B} \citep{qwen_qwen35_2026}; Nemotron 3 Nano, \texttt{nvidia/NVIDIA-Nemotron-3-Nano-30B-A3B-BF16} \citep{nvidia_nvidia_2025}; GPT-5.6 Luna; GPT-5.4 Mini, \texttt{gpt-5.4-mini}; and Gemini 3.6 Flash, \texttt{gemini-3.6-flash}. For the Gemma, Qwen, and Nemotron models, thinking mode is switched on and off through the chat template; Gemini 3.6 Flash, whose thinking cannot be switched off completely, uses the \texttt{minimal} thinking level in non-thinking mode and \texttt{medium} in thinking mode. GPT-5.6 Luna is run as \texttt{gpt-5.6-luna-none} in non-thinking mode and \texttt{gpt-5.6-luna-high} in thinking mode, and GPT-5.4 Mini with reasoning effort \texttt{none} and \texttt{medium}. In thinking mode, the thinking budget is 512 tokens per set of examples for posterior prediction and 2048 tokens for hypothesis evaluation and generation, raised to 4096 tokens for generation in Gemma 4 A4B and Qwen 3.6 A3B, whose thinking otherwise exhausts the budget on \textsc{Bigelow16} stimulus sets.

Every prompt begins with the same description of the integer domain: ``A computer program takes a number from 1 to $d$ and says `yes' or `no' depending on whether the number satisfies a single easy-to-state concept.'' The prompt conditions differ only in the text that frames the observed examples. The \emph{default} prompt adds nothing. The \emph{strong} prompt adds ``These numbers were drawn uniformly at random from all the numbers (1 to $d$) that the program accepts.'' The \emph{weak} prompt adds ``These are known positive examples, but make no assumption about how they were selected.'' The \emph{explicit} prompt adds no sampling text but shows the candidate hypothesis list of Appendix~\ref{sec:appendix_candidates}, introduced by ``The concept is one of the following types:'' and closed by ``Or it could be something else entirely.''; for \textsc{Bigelow16}, it also notes that the listed concepts may appear with linear or exponential transformations.

\section{Stimulus sets and sequential presentation}
\label{sec:appendix_stimuli}

A stimulus set is the full set of unique numbers configured for one run, and the model sees them one at a time. A four-number stimulus set such as $\{16, 8, 2, 64\}$ therefore yields four nested sets of observed examples, $\{16\}$, $\{16, 8\}$, $\{16, 8, 2\}$, and $\{16, 8, 2, 64\}$, each adding one example to the previous one, and a one-number stimulus set yields one. Below, each such set of examples corresponds to one combination of stimulus set and number of examples. \textsc{Tenenbaum99} contains eight stimulus sets: two singletons, three rule-like sets, and three similarity-like sets, which expand into 26 sets of examples per domain (2 singleton, 12 rule-like, and 12 similarity-like). \textsc{Bigelow16} contains 255 stimulus sets, which give 636 sets of examples in the original domain under the same four-example limit. A structural classification labels arithmetic progressions and sets sharing a common factor as rule-like and tight clusters spanning at most 10 as similarity-like, which yields 55 singleton, 289 rule-like, 14 similarity-like, and 278 unclassified sets of examples.

\section{Hypothesis space and candidate hypothesis list}
\label{sec:appendix_candidates}

For a dataset $t$, domain size $d$, and observed examples $X$, let $\mathcal{R}_{t,d}$ denote the fixed rule hypotheses for that dataset and domain, and let $\mathcal{I}^{\mathrm{nat}}_d$ denote the grid interval hypotheses in $\{1,\ldots,d\}$, with endpoints on a 5-integer grid. The Bayesian model uses the full rule-and-interval hypothesis space
\[
\mathcal{H}=\mathcal{R}_{t,d}\cup\mathcal{I}^{\mathrm{nat}}_d.
\]
For \textsc{Tenenbaum99}, the rule set contains parity, squares, cubes, primes, multiples of 3 through 12, powers of 2 through 10, last-digit rules, and the four $5n+k$ residue classes. Merging hypotheses with identical support and removing very small supports leaves 31 rule hypotheses in the original domain and 32 in the extended domain, for 261 and 892 hypotheses after adding the grid intervals. For \textsc{Bigelow16}, the rule set starts from the rule families of \citet{bigelow_inferring_2016}: parity, squares, cubes, primes, and multiples of 3 through 12. It then applies the transformations $n+1$, $n-1$, $n+2$, $n-2$, $2n$, $3n$, $2n+1$, $3n-1$, $3n+1$, $2^n$, $2^{n+1}$, $2^n+1$, and $2^n-1$; merging hypotheses with identical support leaves 128 rule hypotheses in the original domain, which are grouped back into 15 rule-family labels for prompting, and 358 hypotheses after adding the grid intervals. The prior $p_0(h)$ assigns a fraction $\lambda=0.6667$ of its mass uniformly across rule hypotheses and the rest to grid intervals, whose size prior is an Erlang distribution with scale $\sigma=10$. This hand-set prior defines the Bayesian posterior and is the starting point of the pooled prior (Appendix~\ref{sec:appendix_abfit}).

The candidate hypothesis list shown in hypothesis evaluation is an example-dependent subset $K(X)$ of the same rule set and interval family,
\[
K(X)=\mathcal{R}_{t,d}^{\mathrm{prompt}}\cup
\{I_{10}(X), I_{5}(X), I_{\min\max}(X), I_{\mathrm{all}}\}
\cup\{\mathrm{other}\}.
\]
Here $\mathcal{R}_{t,d}^{\mathrm{prompt}}$ contains the 31 or 32 \textsc{Tenenbaum99} rule labels, or the 15 grouped \textsc{Bigelow16} labels. Among the interval candidates, $I_{10}(X)$ and $I_5(X)$ round the smallest and largest observed examples outward to multiples of 10 and 5, $I_{\min\max}(X)$ spans exactly from the smallest to the largest observed example, and $I_{\mathrm{all}}=\{1,\ldots,d\}$. The residual option ``other'' collects confidence assigned to hypotheses outside these labels. The candidate hypothesis list therefore has 36 entries for \textsc{Tenenbaum99} in the original domain, 37 in the extended domain, and 20 for \textsc{Bigelow16}; for some sets of examples, interval candidates coincide, which reduces the number of unique entries.

\section{Eliciting the three measurements}
\label{sec:appendix_elicitation}

Posterior prediction queries each target integer $y\in D_d$ separately. The model receives the domain description, the prompt-condition text, and the observed examples, followed by the target question ``Would the program say `yes' to the number $y$? Answer only Yes or No.'' The posterior prediction $q(y\mid X)$ is the probability of Yes, renormalized over the Yes and No answers and read from the answer-token log-probabilities.

The backends that serve GPT-5.6 Luna and Gemini 3.6 Flash return no log-probabilities, so for these two models posterior prediction is estimated from repeated responses. For each set of examples, the model answers every integer in $D_d$ in a single response; this request is sampled 10 times, at temperature 0.7 for GPT-5.6 Luna and with the provider's default sampling for Gemini 3.6 Flash, and $q(y\mid X)$ is the fraction of valid responses that answer Yes to $y$. Log-space metrics such as KL divergence and entropy use the Jeffreys-smoothed frequency $(k+0.5)/(n+1)$, where $k$ of $n$ valid responses answer Yes, whereas linear metrics, including the size-principle fit and extension mass, use the raw frequency.

In thinking mode, posterior prediction separates inferring the concept from answering each target. The model first thinks about the observed examples, instructed to infer the concept without answering any target question, and the resulting thinking is reused to score every target. The model thus spends its thinking budget once per set of examples rather than once per target, and target-level estimates stay comparable to those in non-thinking mode.

Hypothesis evaluation shows the candidate hypothesis list $K(X)$ and asks for a confidence in each entry. Hypothesis generation shows no list and asks the model to propose 10 hypotheses, each with a confidence. In pilot runs, longer generation lists mainly added low-confidence hypotheses and left the high-confidence hypotheses on which the analyses depend unchanged.

\section{Pooled prior and size-principle fit}
\label{sec:appendix_abfit}

\paragraph{Why a pooled prior.} The hand-set prior $p_0$ fixes the relative weights of hypothesis families by hand, and these weights are arbitrary. Fitting the prior separately for each source does not remove this arbitrariness, because the strength of the prior and the size principle trade off during fitting, so that for sharply peaked sources the fitted size principle shifts with the prior. We therefore learn one prior jointly from all sources and fit only the size principle per source and number of examples.

\paragraph{Data sources.} The pooled prior is learned from 16 LLM sources and one human source. The LLM sources are the eight models of the panel, each in non-thinking and thinking mode. The human source combines the 255 \textsc{Bigelow16} stimulus sets, which carry dense ratings from about 10 raters per stimulus set and target, with the 8 \textsc{Tenenbaum99} stimulus sets, which carry sparse ratings. Every source contributes posterior prediction under the \emph{default} prompt in the original domain, with the two datasets pooled. LLM inputs are the unique sets formed by the first $n=1,\ldots,4$ examples of each stimulus set, deduplicated within each source; human inputs are the full stimulus sets, with $n$ equal to the set size.

\paragraph{Model.} For observed examples $X$ with $n=|X|$ and a size principle $\beta$, the fitted prediction is
\[
\hat q_{\beta}(y\mid X)=
\sum_{h\in\mathcal{H}} \mathbf{1}[y\in h]\,
\frac{\mathbf{1}[X\subseteq h]\,p(h)\,|h|^{-\beta n}}
{\sum_{h'\in\mathcal{H}}\mathbf{1}[X\subseteq h']\,p(h')\,|h'|^{-\beta n}},
\]
where the pooled prior reshapes the hand-set prior $p_0$ of Appendix~\ref{sec:appendix_candidates},
\[
\log p(h)=\alpha\log p_0(h)+a_{\mathrm{base}(h)}+c_{\mathrm{transform}(h)}+\mathrm{const}.
\]
The base offsets $a$ index rule families (odd, even, multiples of $k$, primes, squares, cubes, powers, and so on), and all interval hypotheses share one base. The transformation offsets $c$ index the \textsc{Bigelow16} rule transformations: identity, $n\pm1$, $n\pm2$, $2n$, $3n$, $2n+1$, $3n\pm1$, $2^n$, $2^{n+1}$, and $2^n\pm1$; \textsc{Tenenbaum99} rules are untransformed. The Bayesian model is the unfitted case, $p_0$ with $\alpha=\beta=1$ and no offsets.

\paragraph{Estimating the pooled prior.} One $\alpha$ and the offsets $a$ and $c$ are estimated jointly over all sources, with one size principle $\beta_{j,n}$ for each source $j$ and number of examples $n$. Let $q_j(y\mid X)$ be the posterior prediction of source $j$, $\mathcal{D}$ the set of all pairs $(j,X)$ in the data, and $\mathcal{Y}_X$ the rated targets for $X$. The objective is the squared error averaged over rated targets and then over pairs,
\[
\mathcal{L}=
\frac{1}{|\mathcal{D}|}\sum_{(j,X)\in\mathcal{D}}
\frac{1}{|\mathcal{Y}_X|}\sum_{y\in\mathcal{Y}_X}
\big(q_j(y\mid X)-\hat q_{\beta_{j,|X|}}(y\mid X)\big)^2.
\] A Gaussian penalty on $\log\beta$ centered at 0 (weight $\kappa=0.01$) anchors each size principle at the Bayesian value of 1, a ridge penalty of $10^{-3}$ regularizes the offsets, and the objective is minimized with L-BFGS. We estimate one pooled prior per prompt condition. The \emph{default} prompt's prior includes the human source; the \emph{strong}, \emph{weak}, and \emph{explicit} priors use the LLM sources only, because human participants never saw those prompts.

\paragraph{Size-principle fits.} Downstream fits fix the pooled prior and estimate only $\beta$, with the same penalty on $\log\beta$. We fit one size principle per LLM source, measurement, prompt condition, and number of examples, always pooling the two datasets. Fits for $n=1,\ldots,4$ use the stimulus sets with at least four examples, truncated to their first $n$ examples; the all-example fit uses each stimulus set with all of its examples. Posterior-prediction fits use the pooled prior of their own prompt condition, whereas hypothesis-evaluation, hypothesis-generation, and human fits use the \emph{default} prompt's pooled prior. Because the prior is shared, size principles are comparable across models, numbers of examples, and measurements, whereas prompt conditions are compared under their own priors. Within each prompt condition, $\alpha$ is a single global constant and is not interpreted for any model.

\paragraph{Estimated prior.} Under the \emph{default} prompt, $\alpha=0.93$; it is 0.74 under the \emph{explicit} prompt and about 0.92 under the \emph{strong} and \emph{weak} prompts. The learned offsets favor parity, multiples of 5, primes, and small shifts, and they disfavor interval hypotheses and compound multiplicative or exponential transformations (Table~\ref{tab:pooled_prior}).

\begin{table}[htbp]
  \centering
  \caption{\textbf{Learned offsets of the pooled prior under the \emph{default} prompt.} Selected base and transformation offsets on the log scale, added to $\alpha\log p_0(h)$ with $\alpha=0.93$. Positive values raise the prior of a hypothesis family relative to the hand-set prior, and negative values lower it.}
  \label{tab:pooled_prior}
  \small
  \begin{tabular}{lr@{\hspace{3em}}lr}
    \toprule
    Base & Offset & Transformation & Offset \\
    \midrule
    Even & $+2.24$ & Identity & $+2.12$ \\
    Odd & $+1.03$ & $n\pm1$ & $\approx+0.9$ \\
    Multiples of 5 & $+0.89$ & $n\pm2$ & $+0.60$ to $+0.75$ \\
    Primes & $+0.85$ & $2n$ & $+0.4$ \\
    Multiples of 3 & $+0.68$ & $3n$, $2n+1$ & $\approx0$ \\
    Powers of 2 & $-0.64$ & $3n\pm1$ & $-0.70$ to $-1.06$ \\
    Cubes & $-0.94$ & $2^n\pm1$ & $\approx-1.2$ \\
    Multiples of 6 & $-1.08$ & & \\
    Interval & $-1.63$ & & \\
    \bottomrule
  \end{tabular}
\end{table}

\section{Choice of prior parameterization}
\label{sec:appendix_prior_comparison}

We chose the pooled prior by comparing nine parameterizations under 4-fold cross-validation on the \emph{default}-prompt sources above. Folds are keyed by input, so a held-out input is held out for every source at once, and the score is the pooled squared Pearson correlation $r^2$ between held-out predictions and data over rated targets. Every variant fits one size principle per source and number of examples, so the variants differ only in the prior (Table~\ref{tab:prior_comparison}).

\begin{table}[htbp]
  \centering
  \caption{\textbf{Held-out fit of alternative prior parameterizations.} Pooled squared Pearson correlation $r^2$ between held-out predictions and data under 4-fold cross-validation keyed by input, averaged over the 16 LLM sources and computed for human judgments. Variants 1--4 keep the shape of the hand-set prior $p_0$; variants 5--9 add the learned offsets $a_{\mathrm{base}(h)}+c_{\mathrm{transform}(h)}$. The selected variant (9, bold, $\star$) matches the best-fitting variants with the fewest parameters.}
  \label{tab:prior_comparison}
  \small
  \begin{adjustbox}{max width=\linewidth}
  \begin{tabular}{clrcc}
    \toprule
    Variant & Prior parameterization & Parameters & LLM mean $r^2$ & Human $r^2$ \\
    \midrule
    1 & $\alpha$ per (source, $n$), hand-set prior & 136 & 0.229 & 0.240 \\
    2 & As 1, stronger $\alpha$ regularization (weight 0.1) & 136 & 0.233 & 0.240 \\
    3 & $\alpha$ per $n$, shared across sources & 72 & 0.232 & 0.240 \\
    4 & $\alpha$ per source, shared across $n$ & 85 & 0.229 & 0.240 \\
    5 & $\alpha$ per (source, $n$) + learned offsets & 182 & 0.278 & 0.306 \\
    6 & Learned offsets + one rank-1 coefficient per (source, $n$) & 228 & 0.285 & 0.316 \\
    7 & $\alpha$ per $n$ + learned offsets & 118 & 0.280 & 0.309 \\
    8 & $\alpha$ per source + learned offsets & 131 & 0.280 & 0.308 \\
    \textbf{9}$^\star$ & \textbf{One global $\boldsymbol{\alpha}$ + learned offsets} & \textbf{115} & \textbf{0.280} & \textbf{0.309} \\
    \bottomrule
  \end{tabular}
  \end{adjustbox}
\end{table}

Pooling $\alpha$ differently while keeping the hand-set prior (variants 1--4) barely changes the held-out fit. Learning the offsets accounts for almost all of the gain (variants 5--9), raising the held-out $r^2$ from 0.229 to 0.280 for the LLM mean and from 0.240 to 0.309 for human judgments. Once the offsets are learned, a single global $\alpha$ (variant 9) matches $\alpha$ per number of examples, per source, or per source and number of examples. The rank-1 variant (6), which lets each source and number of examples scale the learned offsets, fits only slightly better, and its gain comes mostly from a few models. We therefore select variant 9, which leaves the size principle as the only quantity fit per source and number of examples.

\begin{figure}[htbp]
  \centering
  \includegraphics[width=0.9\linewidth]{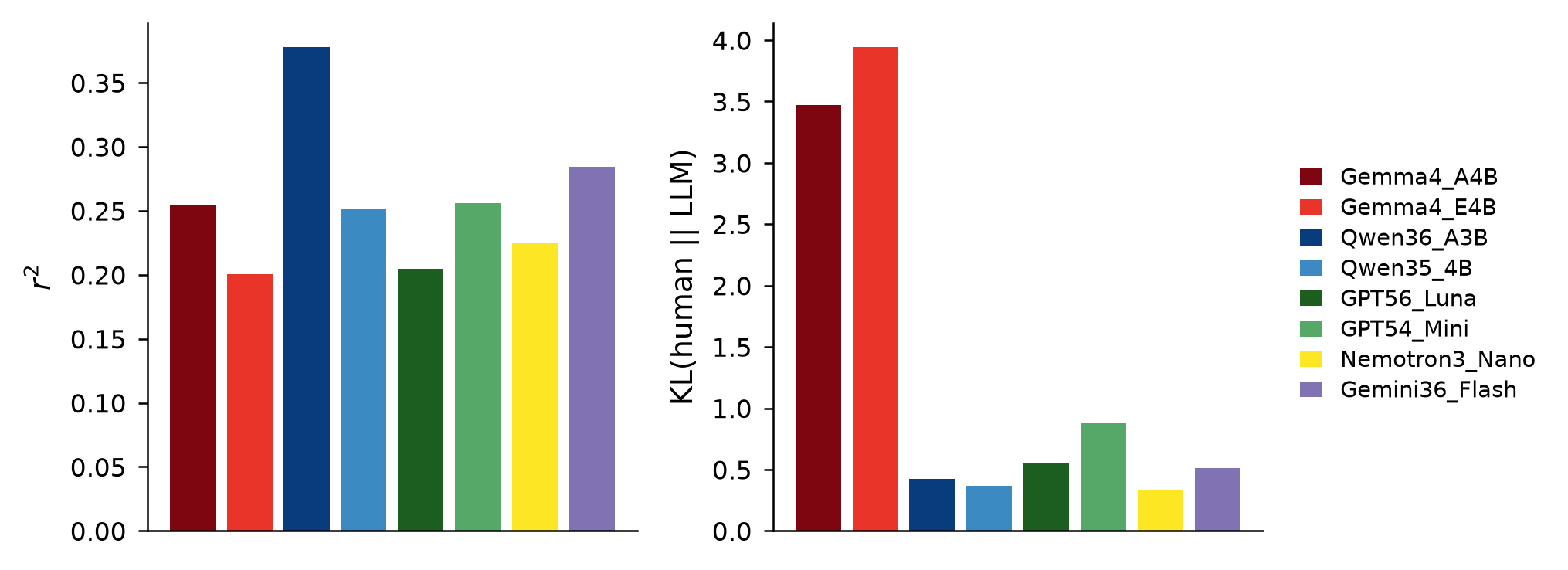}
  \caption{\textbf{Per-model fit quality under the pooled prior and divergence from human judgments.} Left, in-sample $r^2$ of the all-example size-principle fit for each non-thinking model under the \emph{default} prompt's pooled prior. Right, mean KL divergence from human judgments to each model's posterior predictions, KL(human\,$\|$\,LLM), over every human-rated stimulus set under the \emph{default} prompt.}
  \label{fig:supp_fitting_human_kl}
\end{figure}

For human judgments, the noise ceiling is about 0.73 (split-half reliability, Spearman--Brown corrected), and freely reweighting the hypothesis space for each stimulus set reaches 0.64 in sample. In-sample all-example fits are optimistic, because the prior was learned on the same data, but show the same improvement over the hand-set prior: $r^2$ rises from 0.251 to 0.315 for human judgments and from 0.233 to 0.275 for the LLM mean, improving for 13 of the 16 LLM sources, with slight decreases for non-thinking GPT-5.6 Luna (0.214 to 0.205), Gemma 4 E4B (0.204 to 0.200), and Qwen 3.5 4B (0.262 to 0.251). Per-model in-sample fit quality is shown in Fig.~\ref{fig:supp_fitting_human_kl}. Other choices did not improve held-out fit: stronger or weaker regularization, bounds on $\alpha$, rule-mass and interval-size priors, a mixture of weak and strong sampling, sublinear scaling with the number of examples, a lapse or response-noise term, a cross-entropy loss, grouping by stimulus type (rule-like or similarity-like) or dataset, and hierarchical parameters per stimulus set, which describe the fitted stimulus sets better but do not generalize to new ones. Because the pooled prior is learned from the same data that the size-principle fits describe, held-out $r^2$ is the primary fit measure and in-sample values are labeled as such; even the selected prior stays well below the human noise ceiling and below free reweighting per stimulus set, so part of the structure in human and LLM judgments lies outside this prior family.

\section{Projection and top-1 metrics}
\label{sec:appendix_hypmetrics}

Evaluation and generation return hypotheses as text labels, each with a confidence. Before projection, these confidences are normalized within each response so that the retained positive values sum to one; we call the normalized confidences weights. For generation, labels that map to the same rule or interval support are first collapsed, keeping the larger confidence.

To compare the measurements, we map weighted labels into predictive functions over $D_d$. Let $r\in\{\mathrm{eval},\mathrm{gen}\}$ index evaluation and generation, let $L_r(X)$ be the weighted labels returned for examples $X$, a weighted subset of $K(X)$ for evaluation and the model-proposed labels for generation, let $w_r(\ell\mid X)$ be the weight of label $\ell$, and let $s_\ell(y)$ be its support when it matches a rule or interval: $s_\ell(y)=\mathbf{1}[y\in h]$ for a label that matches hypothesis $h$, and, for a grouped \textsc{Bigelow16} label, the fraction of its transformed rules containing $X$ that also contain $y$ (or of all its rules when none contains $X$). The projected prediction is
\[
\tilde q_r(y\mid X)=\sum_{\ell\in L_r(X)} w_r(\ell\mid X)\,s_\ell(y).
\]
Labels that match no rule or interval, including ``other'', are excluded from the projection, because projecting them would require an external judge. For all three measurements, KL divergence and predictive entropy are computed after normalizing the prediction over $D_d$. KL divergence from the Bayesian posterior is $\mathrm{KL}(p\,\|\,q)$, with the Bayesian prediction $p$ first and $0.001$ uniform mass mixed into the LLM prediction $q$ (or $\tilde q_r$), so that a target the LLM rules out costs a bounded penalty; predictive entropy is the Shannon entropy in nats.

The top-1 hypothesis $\ell^\star$ is the highest-weight label that matches a rule or interval, so evaluation skips any ``other'' entry ranked above it. Its normalized confidence is its weight, its support fraction is $\sum_{y}s_{\ell^\star}(y)/d$, and its accuracy is one when $s_{\ell^\star}(x)=1$ for every $x\in X$ and zero otherwise. Rule proportion is averaged over all answered sets of examples: it is one when $\ell^\star$ is a rule and zero when it is an interval or when the response contains no label that matches a rule or interval, so such responses count as non-rule. Top-1 summaries in the main text pool the two datasets in the original domain; human values are computed in the same way from participants' written descriptions of \textsc{Bigelow16} sets.

For rule AUC, each stimulus set is scored with all of its examples by its top-3 rule share, the rule weight divided by the rule-plus-interval weight among its three most confident matched hypotheses. Rule AUC is the probability that a rule-like stimulus set receives a higher score than a similarity-like stimulus set from the same dataset, with ties counted as one half, so 0.5 is chance. The human rule AUC scores each \textsc{Bigelow16} set by the proportion of rule descriptions among its rule and interval descriptions.

\section{Domain-extension metrics}
\label{sec:appendix_domain}

In the domain-extension analysis, the observed examples remain in the original domain while the prompt states the extended domain, so posterior prediction $q(y\mid X)$ covers $\{1,\ldots,200\}$. The extension mass is the share of the normalized prediction that falls in the new range,
\[
M_{\mathrm{ext}}=\frac{\sum_{y=101}^{200}q(y\mid X)}{\sum_{y=1}^{200}q(y\mid X)},
\]
so a model's overall rate of Yes answers cancels. For evaluation and generation, the extension mass is computed in the same way from the projected prediction $\tilde q_r(y\mid X)$ over the extended domain (Appendix~\ref{sec:appendix_hypmetrics}). The analysis uses \textsc{Tenenbaum99} stimulus sets.

To summarize how extension mass changes with the number of examples, we compute for each stimulus set the decline $M_{\mathrm{ext}}(1)-\tfrac{1}{3}\sum_{n=2}^{4}M_{\mathrm{ext}}(n)$, where $M_{\mathrm{ext}}(n)$ is the extension mass after the first $n$ examples, average it within rule-like and within similarity-like stimulus sets, and report the similarity-like minus rule-like difference for each model and measurement. Stimulus sets whose prediction is zero for every target at some number of examples have undefined extension mass and are excluded before averaging.

For rule-consistent extension in the extended domain, we average $q(y\mid X)$ over targets in the new range that satisfy the rule implied by the examples and, separately, over targets that do not. This separates extension to rule-consistent targets from mass spread across the whole new range.

\section{Supplementary figures for the eight-model panel}
\label{sec:appendix_supp_figs}

\begin{figure}[htbp]
  \centering
  \includegraphics[width=0.85\linewidth]{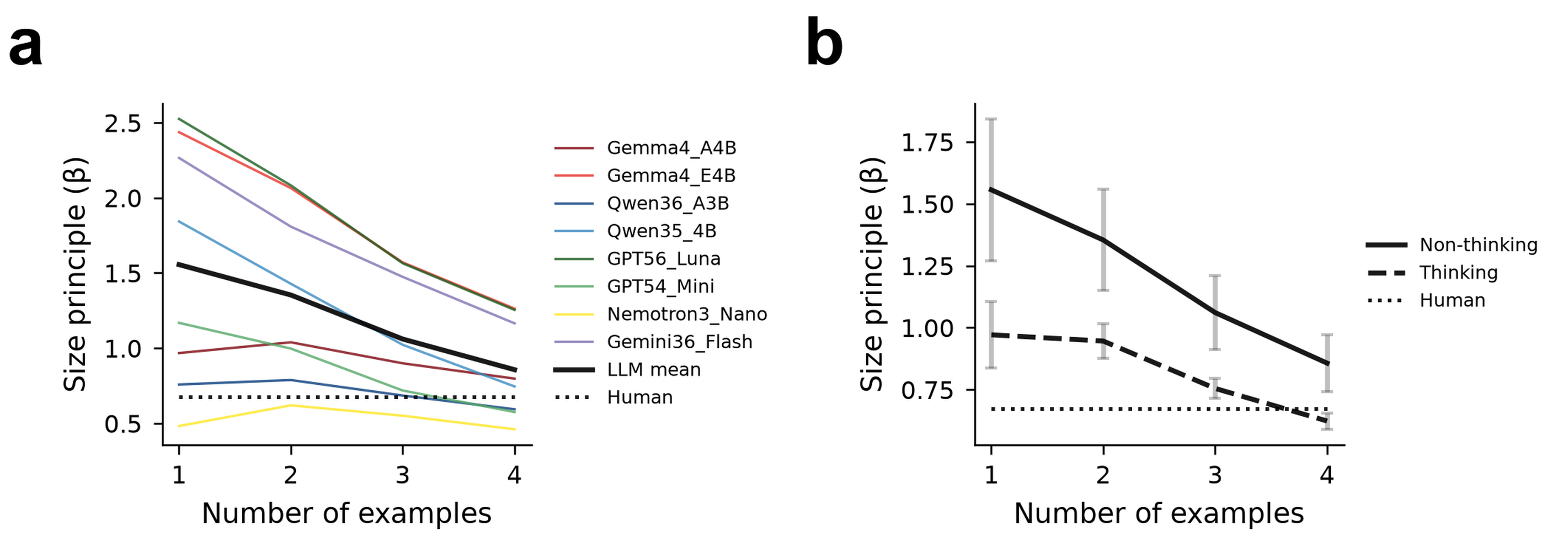}
  \caption{\textbf{The posterior-prediction size principle weakens as examples accumulate, and thinking weakens it further.} \textbf{a}, Size principle over one to four examples for posterior prediction in the original domain under the \emph{default} prompt's pooled prior, fit to stimulus sets with at least four examples and pooled over both datasets. Each colored line is one non-thinking model, the thick black line is the mean across models, and the dotted line marks the human value. \textbf{b}, Mean size principle over paired models in non-thinking and thinking mode; error bars show standard errors across models, and the dotted line marks the human value.}
  \label{fig:supp_alpha_beta_thinking}
\end{figure}

\begin{figure}[htbp]
  \centering
  \includegraphics[width=0.75\linewidth]{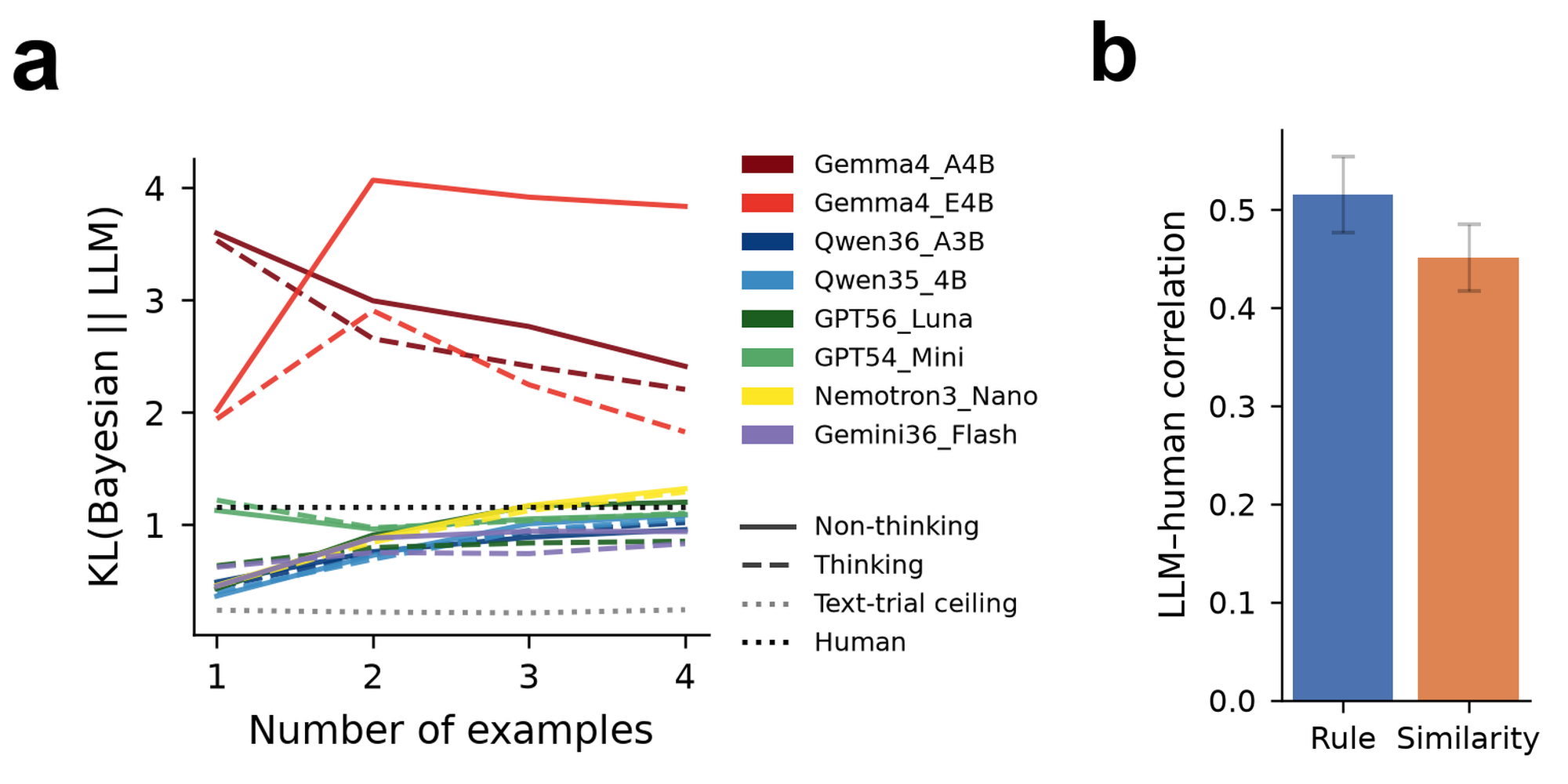}
  \caption{\textbf{Model-level posterior-prediction divergence from the Bayesian posterior and correlation with human judgments.} \textbf{a}, KL divergence from the Bayesian posterior over one to four examples for each model, with solid lines for non-thinking and dashed lines for thinking mode. The dotted black line marks the human value, and the dotted gray line marks the KL that the Bayesian posterior itself would receive if measured with the same 10 repeated responses used for GPT-5.6 Luna and Gemini 3.6 Flash (Appendix~\ref{sec:appendix_elicitation}). \textbf{b}, Correlation between LLM and human posterior predictions on rule-like and similarity-like stimulus sets, pooled over both datasets; error bars show standard errors across models.}
  \label{fig:supp_fit_human}
\end{figure}

\begin{figure}[htbp]
  \centering
  \includegraphics[width=\linewidth]{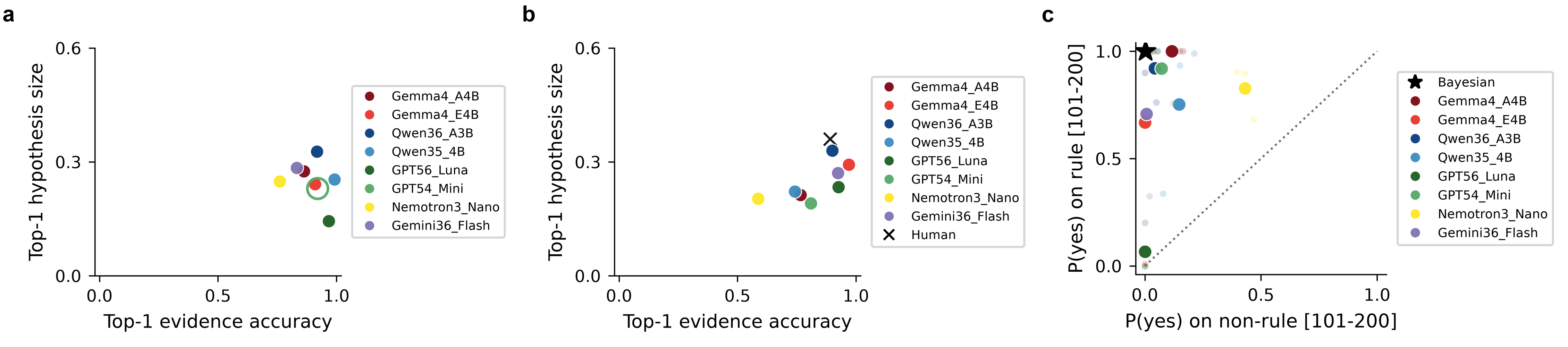}
  \caption{\textbf{Model-level top-1 hypotheses and rule-consistent extension in the extended domain.} \textbf{a,b}, Top-1 support fraction versus top-1 accuracy for hypothesis evaluation (\textbf{a}) and hypothesis generation (\textbf{b}) in the original domain, with one point per non-thinking model. \textbf{c}, Posterior-prediction discrimination between rule-consistent and rule-inconsistent numbers in $101..200$ for rule-like stimulus sets in the extended domain. Transparent points show stimulus sets, larger points show model averages, the black star marks the Bayesian model, and the dotted diagonal marks equal acceptance of both kinds of target.}
  \label{fig:supp_scatter_rule}
\end{figure}

\begin{figure}[htbp]
  \centering
  \includegraphics[width=0.85\linewidth]{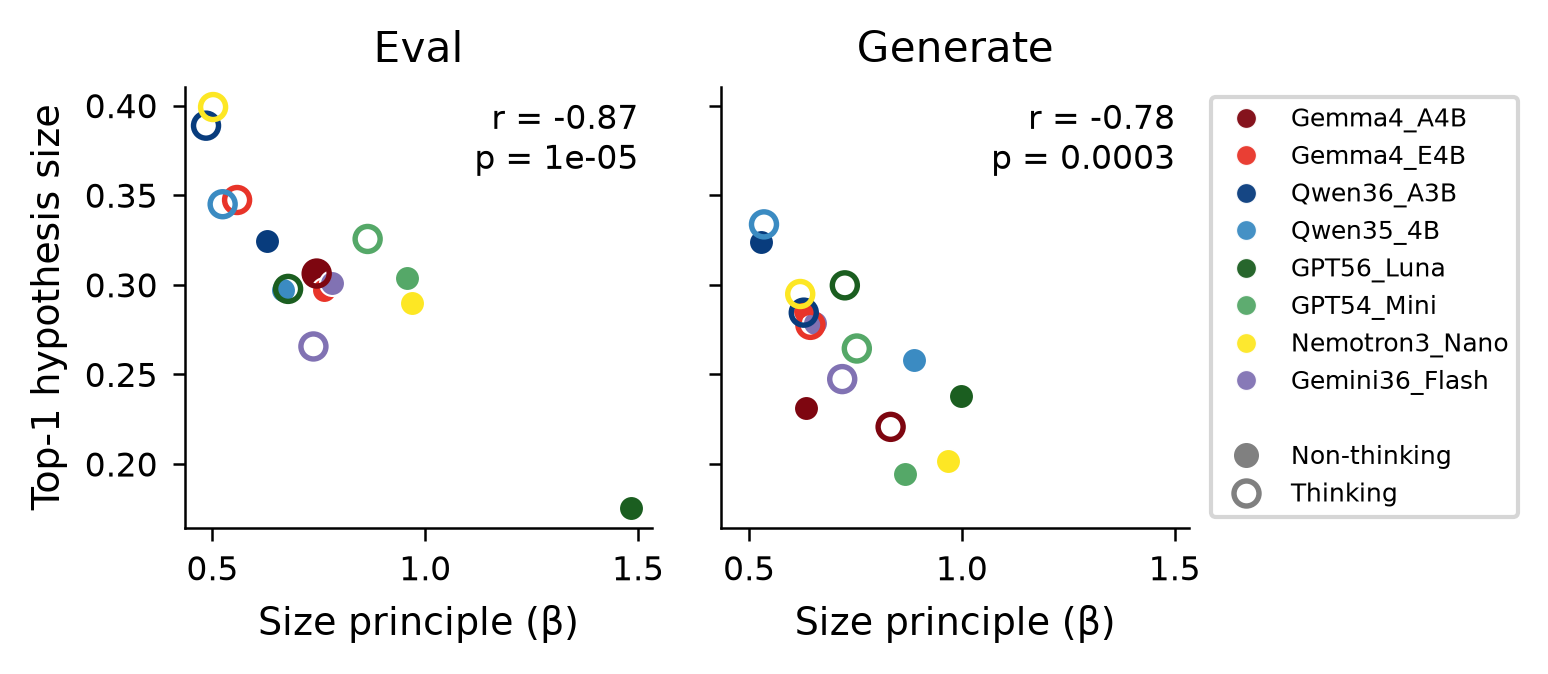}
  \caption{\textbf{Models with a stronger size principle choose narrower top-1 hypotheses.} Mean top-1 support fraction (Hypothesis size), computed with all examples of each stimulus set, versus the all-example size-principle fit for hypothesis evaluation (Eval) and hypothesis generation (Generate) in the original domain, with one point per model; filled points show non-thinking mode and open points thinking mode. Each panel reports the Pearson correlation across points.}
  \label{fig:supp_beta_size}
\end{figure}

\section{Results in the stronger model group}
\label{sec:appendix_best}

\begin{figure}[htbp]
  \centering
  \includegraphics[width=0.85\linewidth]{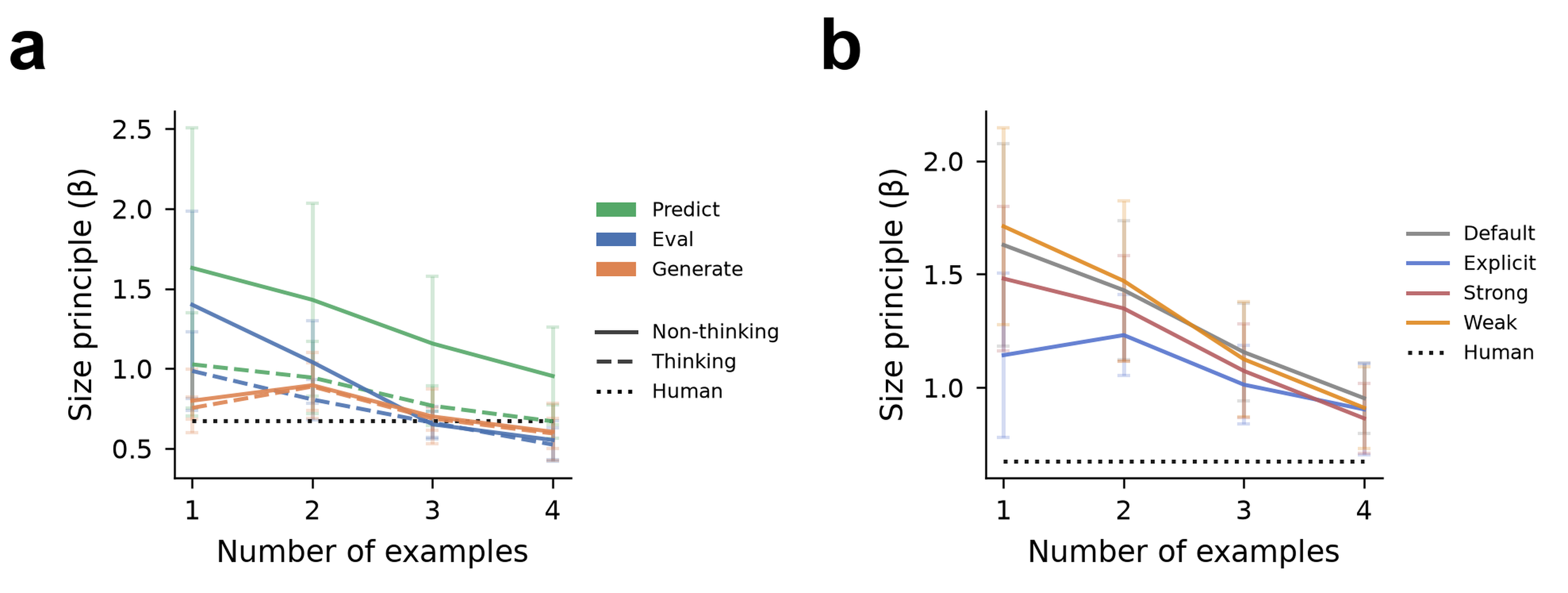}
  \caption{\textbf{Size principle over examples in the stronger model group.} \textbf{a}, Size principle for posterior prediction, hypothesis evaluation, and hypothesis generation, as in Fig.~\ref{fig:behaviors}a, with thinking mode added as dashed lines. \textbf{b}, Posterior-prediction size principle for each prompt condition, each fit under its own pooled prior, as in Fig.~\ref{fig:condition}a.}
  \label{fig:supp_best_alpha_beta_ratio}
\end{figure}

\begin{figure}[htbp]
  \centering
  \includegraphics[width=\linewidth]{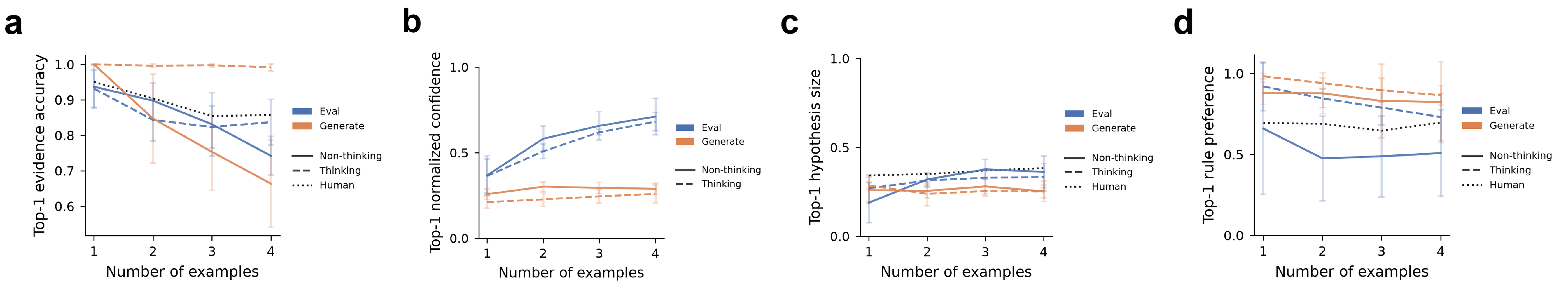}
  \caption{\textbf{Top-1 hypothesis trajectories in the stronger model group.} Top-1 accuracy (\textbf{a}), normalized confidence (\textbf{b}), support fraction (\textbf{c}), and rule proportion (\textbf{d}) for hypothesis evaluation and generation, as in Fig.~\ref{fig:gen_eval_update}.}
  \label{fig:supp_best_update}
\end{figure}

\begin{figure}[htbp]
  \centering
  \includegraphics[width=\linewidth]{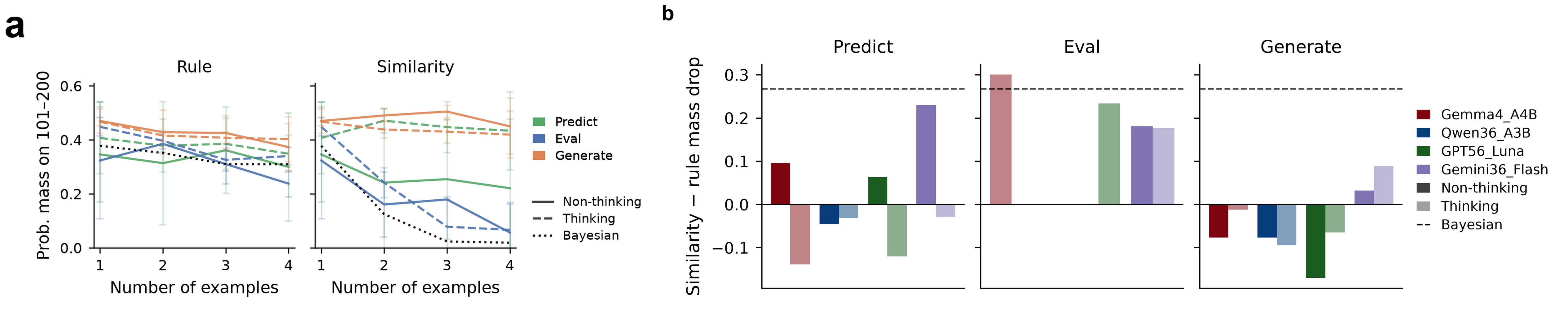}
  \caption{\textbf{Domain extension in the stronger model group.} \textbf{a}, Mass assigned to $101..200$ for posterior prediction, hypothesis evaluation, and hypothesis generation on rule-like and similarity-like stimulus sets as the number of examples grows, as in Fig.~\ref{fig:extension}a. \textbf{b}, Difference between the decline of extension mass on similarity-like and on rule-like stimulus sets for each model and measurement, as in Fig.~\ref{fig:extension}b.}
  \label{fig:supp_best_extension}
\end{figure}

\begin{figure}[htbp]
  \centering
  \includegraphics[width=\linewidth]{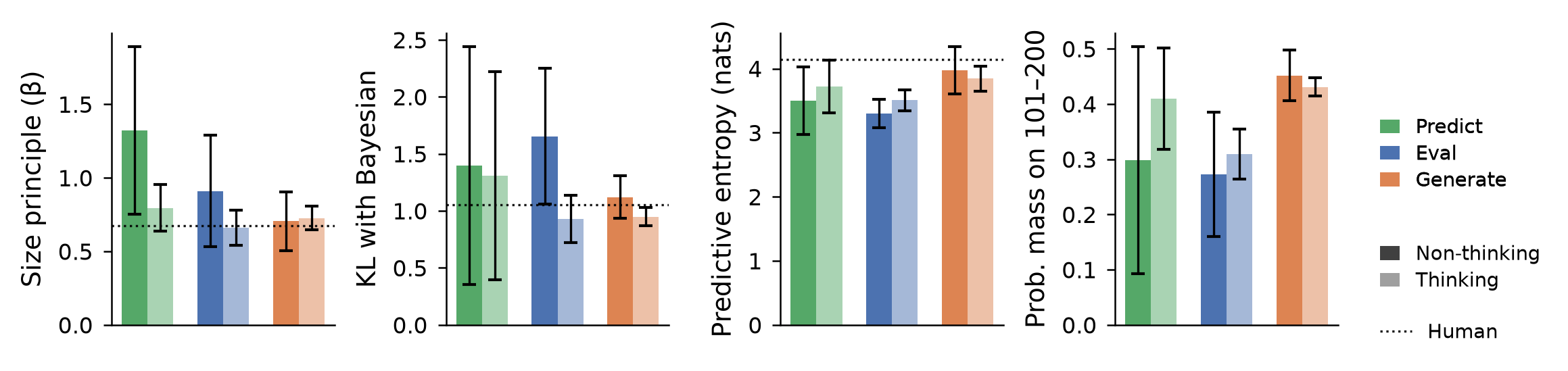}
  \caption{\textbf{Measurement summaries in the stronger model group.} Size principle, KL divergence from the Bayesian posterior, predictive entropy, and mass assigned to $101..200$ in the extended domain for posterior prediction (Predict), hypothesis evaluation (Eval), and hypothesis generation (Generate); the KL and entropy panels correspond to Fig.~\ref{fig:behaviors}b,c.}
  \label{fig:supp_best_condition_task}
\end{figure}

\begin{figure}[htbp]
  \centering
  \includegraphics[width=\linewidth]{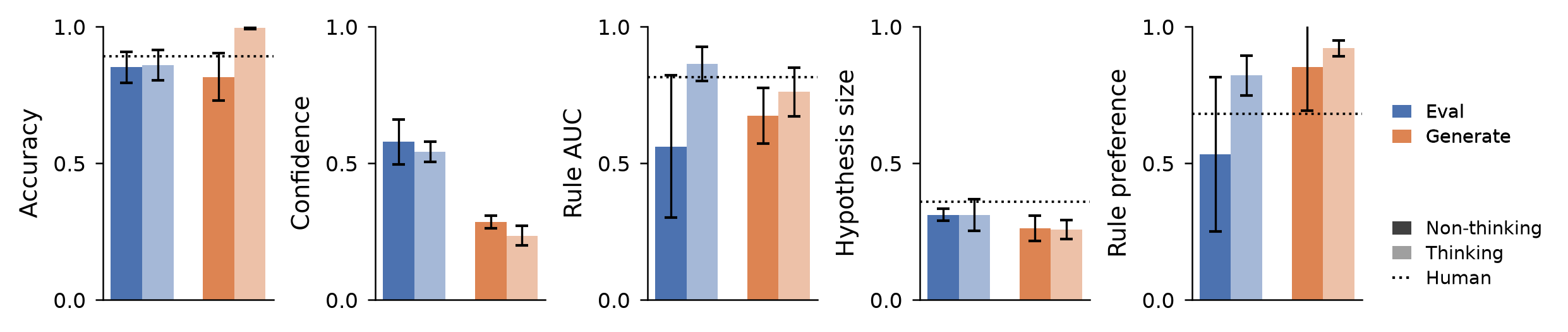}
  \caption{\textbf{Top-1 hypothesis summaries in the stronger model group.} Top-1 accuracy, normalized confidence, rule AUC, support fraction (Hypothesis size), and rule proportion (Rule pref.) for hypothesis evaluation and generation, as in Fig.~\ref{fig:eval_gen_gap}.}
  \label{fig:supp_best_condition_eval_gen}
\end{figure}

\begin{figure}[htbp]
  \centering
  \includegraphics[width=\linewidth]{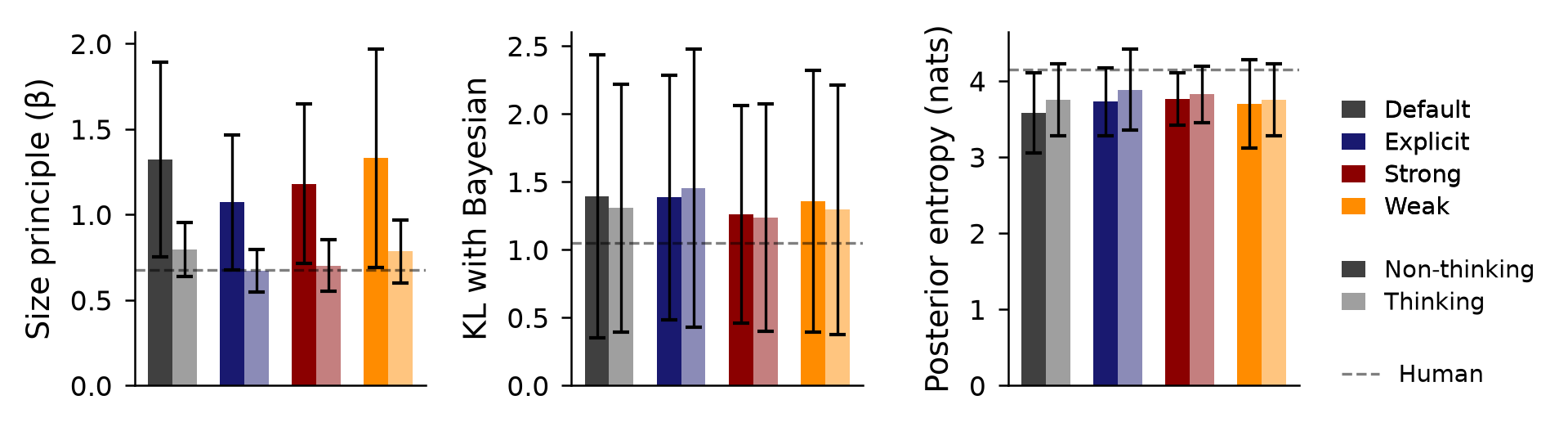}
  \caption{\textbf{Prompt-condition summaries in the stronger model group.} Size principle, KL divergence from the Bayesian posterior, and predictive entropy for each prompt condition in non-thinking and thinking mode, as in Fig.~\ref{fig:condition}b.}
  \label{fig:supp_best_condition}
\end{figure}

Figures~\ref{fig:supp_best_alpha_beta_ratio}--\ref{fig:supp_best_condition} repeat the main-text analyses in the stronger model group: Gemma 4 A4B, Qwen 3.6 A3B, GPT-5.6 Luna, and Gemini 3.6 Flash. Line styles, bar shading, reference lines, and error bars follow the corresponding main-text figures.

\clearpage
\section*{NeurIPS Paper Checklist}

\begin{enumerate}

\item {\bf Claims}
    \item[] Question: Do the main claims made in the abstract and introduction accurately reflect the paper's contributions and scope?
    \item[] Answer: \answerYes{} 
    \item[] Justification: \textcolor{blue}{The abstract and Introduction state the paper's scope: a controlled number-game evaluation of LLM posterior prediction, hypothesis evaluation, and hypothesis generation against a Bayesian model and human judgments. The claims are tied to the five Results figures, and the Discussion and limitations section explicitly bounds the conclusions to this task family and single-run analysis.}

\item {\bf Limitations}
    \item[] Question: Does the paper discuss the limitations of the work performed by the authors?
    \item[] Answer: \answerYes{} 
    \item[] Justification: \textcolor{blue}{The Discussion and limitations section states the main limitations, including the single-run, single-seed design, the narrow task family, the use of only \textsc{Tenenbaum99} in the domain-extension analysis, the asymmetric design of the evaluation--generation comparison, and the dependence of the size-principle fit on the pooled prior and hypothesis space; the limitations of the pooled prior are detailed in the Appendix.}

\item {\bf Theory assumptions and proofs}
    \item[] Question: For each theoretical result, does the paper provide the full set of assumptions and a complete (and correct) proof?
    \item[] Answer: \answerNA{} 
    \item[] Justification: \textcolor{gray}{The paper defines a Bayesian reference model and fitting objectives, but it does not claim new theoretical results requiring formal proofs.}

    \item {\bf Experimental result reproducibility}
    \item[] Question: Does the paper fully disclose all the information needed to reproduce the main experimental results of the paper to the extent that it affects the main claims and/or conclusions of the paper (regardless of whether the code and data are provided or not)?
    \item[] Answer: \answerYes{} 
    \item[] Justification: \textcolor{blue}{The Methods and Appendix define the datasets, domains, sequential example protocol, three measurements of the posterior, prompt conditions, size-principle fit, cross-measurement projection metrics, and domain-extension metrics. The associated experiment repository documents the default runner, configuration files, cache layout, and analysis outputs needed to reproduce the reported figures.}

\item {\bf Open access to data and code}
    \item[] Question: Does the paper provide open access to the data and code, with sufficient instructions to faithfully reproduce the main experimental results, as described in supplemental material?
    \item[] Answer: \answerYes{} 
    \item[] Justification: \textcolor{blue}{We provide open access to the experiment code, data-processing scripts, cached outputs needed for analysis, and figure-generation pipeline in the associated repository, with instructions for reproducing the main results.}

\item {\bf Experimental setting/details}
    \item[] Question: Does the paper specify all the training and test details (e.g., data splits, hyperparameters, how they were chosen, type of optimizer) necessary to understand the results?
    \item[] Answer: \answerYes{} 
    \item[] Justification: \textcolor{blue}{The Methods and Appendix specify the datasets, domains, sequential example protocol, three measurements, prompt conditions, candidate hypothesis list, fitted metrics, and domain-extension metrics. The study evaluates pretrained models rather than training new models, so optimizer and training hyperparameters are not applicable.}

\item {\bf Experiment statistical significance}
    \item[] Question: Does the paper report error bars suitably and correctly defined or other appropriate information about the statistical significance of the experiments?
    \item[] Answer: \answerYes{} 
    \item[] Justification: \textcolor{blue}{The aggregate figures report 95\% confidence intervals wherever model-level summaries are averaged. We do not conduct separate statistical significance tests, and the Discussion and limitations section states that the error bars reflect variation across models or sets of examples rather than repeated stochastic runs.}

\item {\bf Experiments compute resources}
    \item[] Question: For each experiment, does the paper provide sufficient information on the computer resources (type of compute workers, memory, time of execution) needed to reproduce the experiments?
    \item[] Answer: \answerYes{} 
    \item[] Justification: \textcolor{blue}{A subset of local model runs used H100 80GB GPUs; local non-thinking runs typically required 6--12 hours per model, and local thinking runs typically required 24--36 hours per model. Most hosted-model experiments used the Tinker API, with additional OpenAI API runs; total compute is approximately proportional to the number of model-condition runs, while hosted API runs report API usage rather than provider-side worker details.}

\item {\bf Code of ethics}
    \item[] Question: Does the research conducted in the paper conform, in every respect, with the NeurIPS Code of Ethics \url{https://neurips.cc/public/EthicsGuidelines}?
    \item[] Answer: \answerYes{} 
    \item[] Justification: \textcolor{blue}{To the best of our knowledge, the work complies with the NeurIPS Code of Ethics: it analyzes existing pretrained language models on synthetic and previously published number-game stimuli, without collecting private data or deploying a system that affects users.}

\item {\bf Broader impacts}
    \item[] Question: Does the paper discuss both potential positive societal impacts and negative societal impacts of the work performed?
    \item[] Answer: \answerYes{} 
    \item[] Justification: \textcolor{blue}{The Introduction and Discussion motivate the positive impact of better diagnostics for scientific and agentic LLM use, while the Results and Discussion emphasize the negative implication that fluent hypothesis generation can mask incoherent posterior use. The work is foundational and does not introduce a deployed system.}

\item {\bf Safeguards}
    \item[] Question: Does the paper describe safeguards that have been put in place for responsible release of data or models that have a high risk for misuse (e.g., pre-trained language models, image generators, or scraped datasets)?
    \item[] Answer: \answerNA{} 
    \item[] Justification: \textcolor{gray}{The paper does not release a new pretrained model, high-risk scraped dataset, or other asset requiring controlled access. It analyzes existing language models on synthetic and previously published number-game stimuli.}

\item {\bf Licenses for existing assets}
    \item[] Question: Are the creators or original owners of assets (e.g., code, data, models), used in the paper, properly credited and are the license and terms of use explicitly mentioned and properly respected?
    \item[] Answer: \answerYes{} 
    \item[] Justification: \textcolor{blue}{We credit the original number-game sources and use their public releases under the stated terms: the Tenenbaum-style human-rating CSVs are from the GPLv3-licensed \texttt{humanlike\_fewshot\_learning} GitHub repository, and the Bigelow16 number-game dataset is released publicly under Creative Commons Attribution 4.0 (CC BY 4.0). Pretrained model assets are used under the licenses and provider terms listed by their original GitHub, Hugging Face, or API-provider pages.}

\item {\bf New assets}
    \item[] Question: Are new assets introduced in the paper well documented and is the documentation provided alongside the assets?
    \item[] Answer: \answerYes{} 
    \item[] Justification: \textcolor{blue}{We release the experiment code, cached analysis artifacts, and figure-generation pipeline with repository documentation. We do not release a new pretrained model or a newly collected human-subject dataset.}

\item {\bf Crowdsourcing and research with human subjects}
    \item[] Question: For crowdsourcing experiments and research with human subjects, does the paper include the full text of instructions given to participants and screenshots, if applicable, as well as details about compensation (if any)?
    \item[] Answer: \answerNA{} 
    \item[] Justification: \textcolor{gray}{The paper does not collect new human-subject data or use crowdsourcing. Human data appear only as previously published aggregate human judgments and descriptions from the cited number-game literature.}

\item {\bf Institutional review board (IRB) approvals or equivalent for research with human subjects}
    \item[] Question: Does the paper describe potential risks incurred by study participants, whether such risks were disclosed to the subjects, and whether Institutional Review Board (IRB) approvals (or an equivalent approval/review based on the requirements of your country or institution) were obtained?
    \item[] Answer: \answerNA{} 
    \item[] Justification: \textcolor{gray}{No new human-subject experiments, user studies, or crowdsourcing are conducted for this paper, so IRB approval or equivalent review is not applicable for the present study.}

\item {\bf Declaration of LLM usage}
    \item[] Question: Does the paper describe the usage of LLMs if it is an important, original, or non-standard component of the core methods in this research? Note that if the LLM is used only for writing, editing, or formatting purposes and does \emph{not} impact the core methodology, scientific rigor, or originality of the research, declaration is not required.
    \item[] Answer: \answerYes{} 
    \item[] Justification: \textcolor{blue}{LLMs are the core experimental objects of study. The Methods and Appendix describe how pretrained LLMs are prompted and evaluated in posterior prediction, hypothesis evaluation, and hypothesis generation modes, including prompt-condition and thinking comparisons.}

\end{enumerate}

\end{document}